%% file: neurips_2025.tex
\documentclass{article}

\usepackage{xspace}
\usepackage[numbers]{natbib}
\usepackage[lighttt]{lmodern}
\usepackage[dandb, final]{neurips_2025}

\usepackage[utf8]{inputenc} 
\usepackage[T1]{fontenc}    
\usepackage{hyperref}       
\usepackage{url}            
\usepackage{booktabs}       
\usepackage{amsfonts}       
\usepackage{nicefrac}       
\usepackage{microtype}      
\usepackage{xcolor}         
\usepackage{graphicx}

\usepackage{amsmath}
\usepackage[capitalize,noabbrev]{cleveref}
\usepackage{enumitem}

\usepackage[table]{xcolor} 
\usepackage{tabularx}   
\usepackage{amssymb}    
\usepackage{pifont}     
\usepackage{ragged2e}   
\usepackage{makecell}
\usepackage[export]{adjustbox}
\usepackage{collcell}
\usepackage{multirow}
\usepackage{wrapfig}
\usepackage{float}   
\usepackage{listings}
\usepackage{algorithm}        
\usepackage{algpseudocode}    
\usepackage{subcaption}       

\definecolor{ForestGreen}{rgb}{0.13, 0.55, 0.13}
\definecolor{DarkRed}{rgb}{0.64, 0.0, 0.0}

\newcommand{\yesmark}{{\leavevmode\color{ForestGreen}$\checkmark$}}
\newcommand{\nomark}{{\leavevmode\color{DarkRed}$\times$}}
\newcommand{\unspecified}{{\leavevmode\color{gray}--}}
\newcommand{\valignmark}[1]{#1}

\newcommand{\stackheader}[1]{\textbf{\begin{tabular}[t]{@{}c@{}}#1\end{tabular}}}

\newcolumntype{L}[1]{>{\RaggedRight\arraybackslash\hspace{0pt}}p{#1}} 
\newcolumntype{C}[1]{>{\Centering\arraybackslash\hspace{0pt}}p{#1}} 
\newcolumntype{Y}{>{\centering\arraybackslash}X}
\newcolumntype{y}{>{\centering\arraybackslash\scriptsize}X}

\newif\ifcomments
\commentstrue 
\input{commands}

\title{Simulating \textit{Viva Voce} Examinations to Evaluate Clinical Reasoning in Large Language Models}

\author{\small
  Christopher Chiu \\
  Georgia Institute of Technology\\
  \texttt{cchiu65@gatech.edu} \\
  \And
  Silviu Pitis \\
  University of Toronto \\
  \texttt{spitis@cs.toronto.edu} \\
  \And
  Mihaela van der Schaar \\
  University of Cambridge \\
  \texttt{mv472@cam.ac.uk} \\
}

\begin{document}

\maketitle

\begin{abstract}
Clinical reasoning in medicine is a hypothesis-driven process where physicians refine diagnoses from limited information through targeted history, physical examination, and diagnostic investigations. In contrast, current medical benchmarks for large language models (LLMs) primarily assess knowledge recall through single-turn questions, where complete clinical information is provided upfront. To address this gap, we introduce VivaBench, a multi-turn benchmark that evaluates sequential clinical reasoning in LLM agents. Our dataset comprises 1152 physician-curated clinical vignettes structured as interactive scenarios that simulate a viva voce examination in medical training, requiring agents to actively probe for relevant findings, select appropriate investigations, and synthesize information across multiple steps to reach a diagnosis. We evaluated several state-of-the-art LLMs and found that while models demonstrate competence in diagnosing conditions within well-described clinical presentations, their performance degrades significantly when required to navigate diagnostic uncertainty. Our analysis identified several failure modes that mirror common issues in clinical practice, including: (1) fixation on initial hypotheses, (2) excessive investigation ordering, (3) premature diagnostic closure, and (4) missing critical conditions. These patterns reveal fundamental limitations in how current LLMs manage uncertainty and gather information sequentially. Through VivaBench, we provide a standardized benchmark for evaluating conversational medical AI systems for real-world clinical decision support. Beyond medical applications, we contribute to the larger corpus of research on agentic AI by demonstrating how sequential reasoning trajectories can diverge in complex decision-making environments. 
\end{abstract}

\section{Introduction}

Many future applications of Large Language Models (LLMs), such as LLM-assisted clinical reasoning and diagnostics, will involve multi-turn, sequential interactions, where an LLM or LLM-based agent will need to actively gather and act on information that only becomes available at inference time. Most current LLM evaluations, however, including those in healthcare \cite{jin2019pubmedqa} and those aspiring to be comprehensive assessments of general intelligence such as ``humanity's last exam'' \cite{phan2025humanity,chollet2019measure}, evaluate LLMs based on single-turn interactions \cite{liang2022holistic,srivastava2022beyond}. While this approach is valuable for gauging foundational knowledge and specific skills like question answering \cite{jin2021disease} or knowledge encoding \cite{singhal2023large}, it falls short of capturing the dynamic, iterative nature of complex real-world problem-solving that is so central in critical domains like healthcare.

A growing body of recent work examines the multi-turn conversational and reasoning capabilities of LLMs \cite{guan2025evaluating,yi2024survey,zhang2025survey}, and observes that even frontier models exhibit significantly lower performance 
\setlength{\columnsep}{1.4\baselineskip}%
{
\parfillskip=0pt
\parskip=0pt
\par}
\begin{wrapfigure}[21]{R}{0.52\textwidth}
    \vspace{-1.3\baselineskip}
        \centering
        \includegraphics[width=\linewidth]{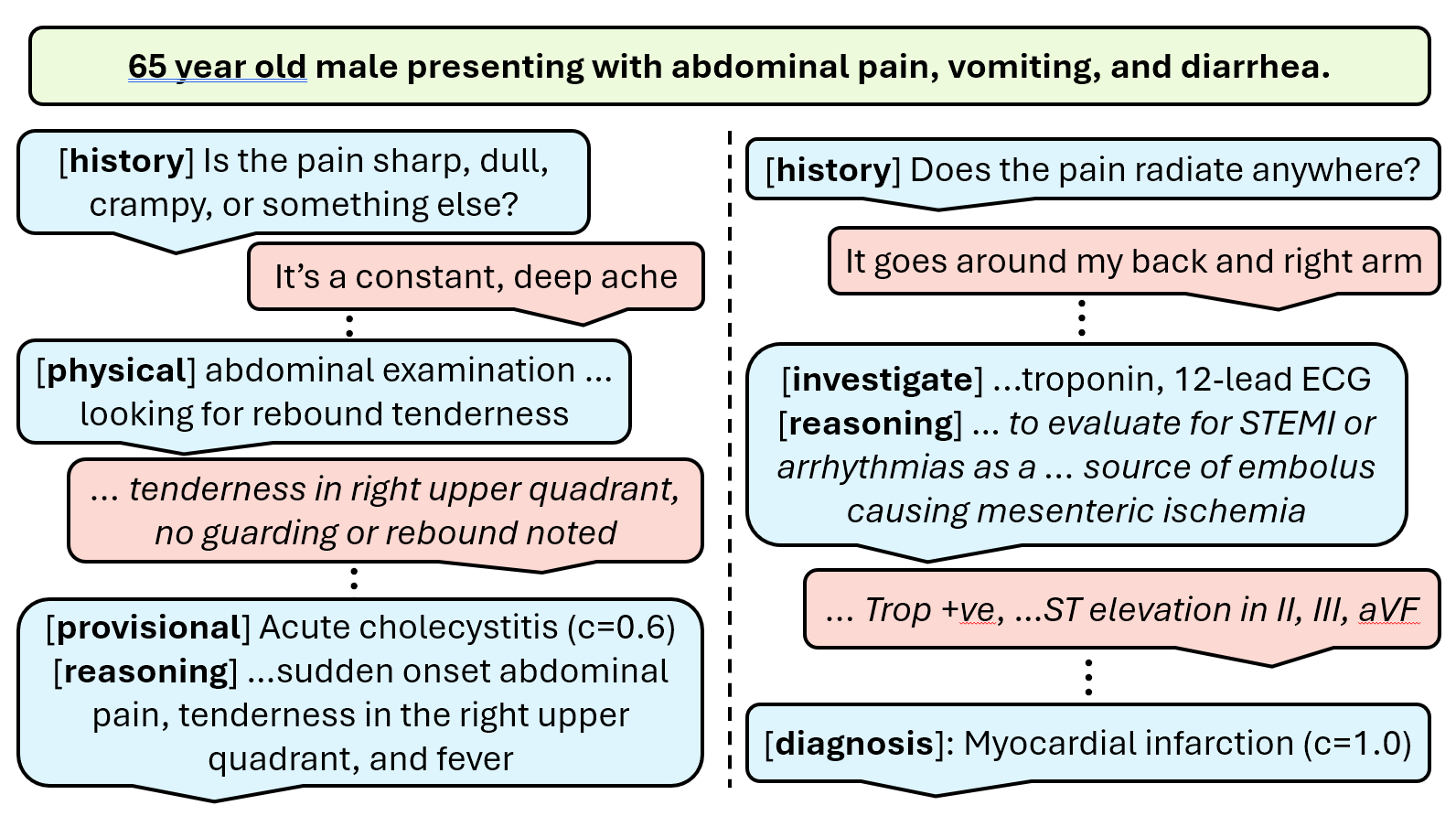}
        \caption{Action and reasoning trace of two evaluated models on our simulated \textit{viva voce} examination. Given the initial scenario (green), agents (blue) are tasked to diagnose the patient (orange), which was simulated by our evaluation framework. \textit{c} indicates confidence of diagnosis. Failure to perform a targeted clinical review (left) could lead to significant ramifications, such as a missed diagnosis of heart attack (correctly diagnosed on the right).}
        \label{fig:example-trace}
\end{wrapfigure}
\noindent in multi-turn settings and suffer from issues of derailment and error propagation when reasoning over multiple steps \cite{kwan2024mt,laban2025llms,sirdeshmukh2025multichallenge}. One such setting, and the subject of this work, is that of clinical decision-making, wherein the derailment of naturally emerging multi-step interactions can have significant consequences for patients.

In clinical practice, the diagnostic journey of a patient can be conceptualized as a Partially Observable Markov Decision Process \cite{steimle2017markov}. Physicians begin with limited, often ambiguous, information (generic findings) and iteratively gather more data through history taking, physical examinations, and diagnostic tests, as is roughly illustrated in \cref{fig:example-trace}. Each new piece of information allows them to update their belief state about the patient's underlying condition, refining their differential diagnosis in a manner akin to Bayesian inference, where the probability of a specific diagnosis given the observed findings, $P(\text{specific diagnosis} | \text{generic findings})$, is continuously updated. This iterative, probabilistic reasoning is fundamental to how clinicians navigate diagnostic uncertainty.

Schubert et al. \cite{schubert2023foundational} have explored the concept of a "diagnostic trajectory" in depth, emphasizing the sequential nature of clinical encounters and the accumulation of evidence over time. Their framework highlights the importance of timely and accurate diagnosis, which relies on the ability to effectively navigate these trajectories. This underscores the need for AI models that are not only knowledgeable but are also capable of robust diagnostic reasoning along these evolving pathways. A critical question then arises: how can we effectively measure and benchmark this sophisticated, sequential reasoning capability in AI systems to ensure they are safe and effective for clinical use?

\paragraph{Viva voce examinations} A viva voce examination is an interactive oral exam used to test the clinical skills of medical students. While it is most commonly used as thesis defense within academia, it has also been used and validated as an examination tool in medical training \cite{abuzied2023structured,daniel2019clinical}. A typical example, as shown in \cref{fig:example-trace}, begins with a simulated patient scenario, and the student physician then queries the patient to answer a series of diagnosis / case management questions. 
The verbal nature of this examination translates nicely, allowing for easy administration to a ``student'' LLM. 

Inspired by this, we propose VivaBench, which operationalizes the viva voce concept for evaluating sequential clinical reasoning capabilities of LLM-based agents. It simulates a multi-turn diagnostic encounter, akin to a medical viva voce examination, where an agent must actively solicit information through history taking, physical examination requests, and ordering of diagnostic investigations to arrive at a diagnosis for a presented clinical case. Our key contributions are:
\begin{itemize}[leftmargin=1.5em]
    \item VivaBench is an {open-source, open-domain dataset} with physician-curated structured clinical vignettes. Each vignette is manually reviewed and structured for interactive probing, requiring agents to navigate diagnostic uncertainty from limited initial information. This interactive format moves beyond static Q and A to assess dynamic information gathering and hypothesis refinement.
    \item We provide an {extensible evaluation framework} for the interactive administration of VivaBench vignettes, which is designed to provide accurate and deterministic outputs for reproducibility.
    \item We evaluate \textbf{6} leading LLMs and systematically {identified critical failure modes}. These patterns mirror common cognitive errors in clinical practice and reveal fundamental limitations in how current LLMs manage uncertainty and synthesize information sequentially in high-stakes scenarios, including: (1) fixation on initial hypotheses (anchoring bias), (2) excessive or inappropriate investigation ordering, (3) premature diagnostic closure, and (4) failure to consider or rule out critical, time-sensitive conditions \cite{omiye2024large, kim2025assessing, busch2024systematic}.
    \item Our work has significant {implications for medical AI development}, providing a tool for assessing conversational AI systems intended for real-world clinical decision support. Beyond medicine, we contribute to a growing body of work on agentic AI by demonstrating how sequential reasoning trajectories can diverge in complex, information-gathering decision-making environments.
\end{itemize}


\section{Related Work}
Early evaluations for language models in medicine, such PubMedQA \citep{jin2019pubmedqa} and  MedQA \citep{jin2021disease}, were based on static question-answering (QA) and assessed knowledge recall from complete information. While valuable for gauging basic clinical knowledge \citep{singhal2023large}, these single-turn evaluations inadequately capture the dynamic, iterative reasoning essential for clinical practice \cite{lateef2021clinical}. High accuracy on such tasks can mask crucial metacognitive deficiencies, such as overconfidence or failure to recognize knowledge gaps, and may not translate to effective real-world decision-making, where LLMs have shown limitations even with full information \citep{hager2024evaluation, omiye2024large, busch2024systematic}. Recent work like MetaMedQA has begun to explore these metacognitive aspects, revealing significant shortcomings in LLMs' ability to handle underspecification and uncertainty both inside and outside medical contexts \cite{ruan2024identifying,griot2025large}. 

Several other multi-turn benchmarks have also emerged to further assess sequential reasoning in medical LLMs. In particular, AI Hospital \citep{fan2024ai} and AgentClinic \citep{schmidgall2024AgentClinicMultimodalAgent} simulate clinical encounters where LLMs must actively gather information. Both benchmarks showed that model performance degrades significantly in multi-turn settings compared to single-turn QA \citep{fan2024ai, kwan2024mt, laban2025llms}, suffering from error propagation and difficulties in maintaining context over extended interactions \citep{guan2025evaluating, sirdeshmukh2025multichallenge}. Concurrently, research into conversational diagnostic AI, such as Google's AMIE \citep{tuConversationalDiagnosticAI2024}, included sophisticated benchmarks with multi-turn medical dialogue. While these benchmarks advance evaluation by incorporating interactivity, they rely on either a human-in-the-loop, or multiple LLM calls within their evaluation chain to account for different agent roles, potentially introducing non-determinism or data leakage, and may not be fully open-source or easily extensible.

\vspace{-\baselineskip}

\begin{table}[!htbp]\footnotesize
\centering
\caption{Comparison of Medical AI Datasets and Frameworks}
\label{tab:medical_ai_stackheader}
\setlength{\tabcolsep}{4pt}

\begin{tabularx}{\columnwidth}{@{} L{1.1in} C{2in} YYYY @{}} 
\toprule
\textbf{\footnotesize Dataset/Framework} & \stackheader{Evaluated\\Capabilities} & \stackheader{Multi-\\turn} & \stackheader{Det.\\Output} & \stackheader{Open\\Source} & \stackheader{Det.\\Criteria} \\
\midrule
QA Datasets \cite{jin2019pubmedqa, jin2021disease} & Knowledge & \valignmark{\nomark} & \valignmark{\unspecified} & \valignmark{\yesmark} & \valignmark{\yesmark} \\
\addlinespace 
AI Hospital \cite{fan2024ai} & Diagnosis & \valignmark{\yesmark} & \valignmark{\nomark} & \valignmark{\yesmark} & \valignmark{\nomark} \\
\addlinespace
AgentClinic \cite{schmidgall2024AgentClinicMultimodalAgent} & Diagnosis, Bias & \valignmark{\yesmark} & \valignmark{\nomark} & \valignmark{\yesmark} & \valignmark{\nomark} \\
\addlinespace
Hager et al. \cite{hager2024evaluation} & Decision making and Diagnosis& \valignmark{\yesmark} & \valignmark{\yesmark} & \valignmark{\nomark} & \valignmark{\yesmark} \\
\addlinespace
Med-PaLM \cite{singhal2023large} & Diagnosis, Conversation Quality & \valignmark{\yesmark} & \valignmark{\yesmark} & \valignmark{\nomark} & \valignmark{\yesmark} \\
\addlinespace
VivaBench \textbf{(ours)} & \makecell{Diagnosis, Bayesian\\reasoning, Information Extraction} & \valignmark{\yesmark} & \valignmark{\yesmark} & \valignmark{\yesmark} & \valignmark{\yesmark} \\
\bottomrule
\end{tabularx}
\par\medskip 
{\vspace{-0.3\baselineskip}\footnotesize 
\yesmark: Yes\quad \nomark: No\quad \unspecified: Not specified or N/A. \hfill
} 
\end{table}

\section{VivaBench}
VivaBench simulates authentic clinical encounters where information must be actively gathered rather than being provided upfront, mirroring real-world diagnostic processes. We implement this by transforming free-text clinical data into a formalized Clinical Case ($\mathcal{C}$) structure. Each Case consists of five primary components that align with standard clinical documentation: History ($\mathcal{H}$), Physical Examination ($\mathcal{P}$), Imaging ($\mathcal{I}$), Laboratory investigations ($\mathcal{L}$), and a ground truth Diagnosis set ($\mathcal{D}$) that includes accepted differential diagnoses ($\mathcal{D}'$).

During evaluation, an agent $\mathcal{A}$ receives an initial clinical stem containing limited background information, and is tasked to progressively work toward a diagnosis with appropriate confidence. The agent can access four distinct action categories corresponding to the structured information components ($\mathcal{H}$, $\mathcal{P}$, $\mathcal{I}$, $\mathcal{L}$). Following each action, an examiner module ($\mathcal{E}$) retrieves and presents the specifically requested clinical information, effectively simulating the progressive information exchange characteristic of clinical viva voce examinations.

The evaluation proceeds through two distinct phases: (i) a \textbf{Review} phase, where the agent interviews the patient ($\mathcal{H}$) and conducts physical examination ($\mathcal{P}$), and (ii) an \textbf{Investigation} phase, where the agent orders laboratory tests or imaging studies ($\mathcal{I}$, $\mathcal{L}$) to refine diagnostic hypotheses. The agent provides a \textbf{provisional diagnosis} with associated confidence level after the \textbf{Review} phase, followed by a \textbf{final diagnosis} after the Investigation phase, at which point the interaction concludes. Both diagnostic assessments are evaluated against the ground truth diagnosis set.

\subsection{Dataset Creation} 

\textbf{Dataset Source and Filtering} A foundational objective of VivaBench is to create an open, accessible benchmark. As such, we sourced clinical vignettes exclusively from publicly available repositories, including MedQA \cite{jin2021disease}, training materials from physician colleges in Australia and the United Kingdom \cite{SampleScenariosFederationa}, and PubMed case reports \cite{2022NcbiPubmedDatasets}. We selected cases based on predefined criteria optimizing for clinical relevance, diagnostic complexity, and educational value. Our selection prioritized cases that: (1) represent common clinical presentations, (2) necessitate multifaceted reasoning across different information types, (3) follow clear diagnostic pathways, and (4) include conditions with significant consequences if misdiagnosed. Appendix \ref{app:dataset_generation} provides a comprehensive description of our selection criteria.

\textbf{Structuring Clinical Data} To parse any free-text clinical information into a structure format, a standardized format was required. We designed an ontology according to standard clinical documentation guidelines\cite{RACGPClinicalGuidelines, ClinicalPracticeGuideline}, and implemented a hierarchical schema for organizing clinical information. Each Clinical Case ($\mathcal{C}$) is partitioned into History ($\mathcal{H}$), Physical Examination ($\mathcal{P}$), Investigations ($\mathcal{I}$), and Laboratory tests ($\mathcal{L}$). History items are further characterized through attributes corresponding to the SOCRATES framework (Site, Onset, Character, Radiation, Associated symptoms, Time course, Exacerbating/relieving factors, Severity) for pain and general symptoms \cite{gregory2019use}, supplemented by standard history categories (past medical history, medications, allergies, family history, social history). Physical examination findings are systematically organized by body systems. Investigations and Imaging are categorized by modality and specimen type, with each diagnostic test result accompanied by appropriate reference ranges where applicable.

To ensure standardization and facilitate information retrieval, each clinical data element is mapped to established medical terminologies (SNOMED-CT for $\mathcal{H}$ and $\mathcal{P}$, LOINC codes for $\mathcal{I}$ and $\mathcal{L}$, and ICD-10 codes for $\mathcal{D}$). This standardization enables precise matching between agent queries and available information. Appendix \ref{app:schema_design} provides a comprehensive description of our dataset schema.

\textbf{Generation Pipeline} Our dataset generation pipeline consists of both automated processing steps to handle volume and human review steps to ensure quality. We first screened for cases in large databases that fit our selection criteria with traditional NLP techniques and LLM assistance. Then, clinical information was extracted from source materials and systematically categorized into respective information components ($\mathcal{H, P, I, L}$). Each component was further subdivided and assigned retrieval keys. Diagnoses were either parsed directly from the original vignettes, or annotated by clinical experts when not explicitly stated. In cases where information was implicit or incomplete, clinicians provided appropriate additions based on standard clinical knowledge to ensure comprehensive case representation. Finally, all structured clinical vignettes underwent physician review to verify information accuracy before inclusion in the dataset. We have made our generation codebase publicly available to enable further case generation from any clinical source material.

\textbf{Dataset Description} Our generation pipeline produced 1,952 cases of parsed structured data. Following human review, 990 cases were retained, all derived from the PubMed database \cite{2022NcbiPubmedDatasets}. Both the complete generated dataset and human-reviewed subset are available on our HuggingFace repository. Each entry includes a unique identifier (\textit{uid}), \textit{source} information, free-text clinical \textit{vignette}, \textit{diagnosis} with \textit{differentials}, and a \textit{clinicalcase} field containing a JSON string representing a Case object compatible with our evaluation framework. Appendix \ref{app:dataset_generation} provides detailed documentation of our dataset generation methodology.

\subsection{Evaluation framework} 

Our evaluation framework provides the code infrastructure to evaluate any conversational AI agent $\mathcal{A}$ against our structured clinical cases. A unit of evaluation in our framework is an Examination, with an Examiner module ($\mathcal{E}$). $\mathcal{E}$ is responsible for the majority of work in our evaluation, including processing information requests, retrieving relevant data from the structured case, and returning information in a natural language format to $\mathcal{A}$. The evaluation workflow proceeds as follows:

\begin{algorithm}
\caption{Clinical Diagnostic Evaluation Process}
\begin{algorithmic}[1]
\State $\mathcal{A}$ is presented with an initial clinical stem, and enters the \textbf{Review} phase of evaluation.
\State $\mathcal{A}$ queries for bedside information using natural language
\State $\mathcal{E}$ processes these queries, and returns relevant information from $\mathcal{H}$ and $\mathcal{P}$
\State Steps 2, 3 are repeated, until $\mathcal{A}$ submits a provisional diagnosis, and proceeds to the \textbf{Investigation} phase of evaluation.
\State $\mathcal{A}$ orders diagnostic investigations, including both imaging and laboratory investigations.
\State $\mathcal{E}$ processes these queries, and returns relevant information from $\mathcal{I}$ and $\mathcal{L}$
\State Steps 5, 6 are repeated, until $\mathcal{A}$ submits a final diagnosis
\State Agent performance is evaluated based on accuracy of provisional and final diagnosis.
\end{algorithmic}
\end{algorithm}

\textbf{Information Retrieval and Parsing} A core technical challenge in our framework is the translation between natural language queries and structured information retrieval. This process involves four sequential steps: receiving the free-text query from the agent, mapping this query to structured keys, retrieving structured information based on matched keys, and transforming this structured information into a natural language response. We provide both deterministic and LLM-based implementations of the mapper and parser components. The deterministic mapper employs cosine similarity embeddings to identify the most relevant keys, complemented by keyword matching and medical entity recognition. In contrast, the LLM-based mapper utilizes few-shot examples to interpret the semantic intent of queries and match them to appropriate information categories and keys. For history and physical examination findings, negative results (absent symptoms or normal examination findings) are explicitly returned when queried. Standardized laboratory values not specifically mentioned in the case are returned as default normal values with appropriate reference ranges, while investigations not available in the case are explicitly noted as ``not available'' to prevent information leakage. To validate our information retrieval system, we measured inter-annotator agreement on query-to-information mapping. We outline our experiments and validation results in the Appendix.

\textbf{Metrics} To evaluate diagnostic performance, we assessed both accuracy and diagnostic reasoning process. We computed top-k exact and approximate accuracies for k=1 through 5, where a prediction was considered an exact match if its ICD-10 code matched a ground truth diagnosis at the appropriate hierarchical level (e.g., a prediction of E78.1 matches ground truth E78), or an approximate match if it matched an accepted differential, shared the top levels of ICD-10 code with any diagnosis, or had high semantic similarity to the correct diagnosis. For cases with multiple ground truth diagnoses, accuracy was counted if any one diagnosis was correctly identified. We also measured model confidence using a weighted score:
{\small
\[S_\textrm{conf} = \sum_{i \in E} c_i + \sum_{i \in A} c_i - \sum_{i \in U} c_i\]
}%
\noindent%
where $c_i$ 
represents the normalized confidence for each prediction belonging to exact matches ($E$), approximate matches ($A$), or unmatched predictions ($U$). This score ranges from -1.0 (high confidence in wrong diagnoses) to 1.0 (high confidence in correct diagnoses).

Additionally, we tracked the relevance of clinical information requested by calculating precision (proportion of requested information that was relevant to the diagnosis: $|K_\textrm{requested} \cap K_\textrm{relevant}|/|K_\textrm{requested}|$) and recall (proportion of all diagnosis-relevant information that was requested: $|K_\textrm{requested} \cap K_\textrm{relevant}|/|K_\textrm{relevant}|$). These metrics were calculated for different information categories (history/physical vs. investigations) and scopes (targeted to matched diagnoses vs. all potential diagnoses). Finally, we measured how diagnoses evolved from provisional to final stages to evaluate how effectively models updated their reasoning when presented with additional information. 

\section{Evaluation}
We evaluated several state-of-the-art foundation models, including Gemini 2.5 pro \cite{GoogleGemini2_5Pro2025}, DeepSeek-R1 \cite{DeepSeekAI2025DeepSeekR1}, o4-mini from OpenAI \cite{OpenAIO3O4miniSystemCard2025}, Llama-4 Maverick \cite{MetaLlama4Maverick2025}, Grok 3 mini beta \cite{xAI_Grok3mini_beta_2025}, and Qwen 3 (235b-a22b) \cite{AlibabaQwen3_235B_A22B_2025}. Each model was tested at temperature 0, and instructed with the task to perform a clinical review and diagnosis on the patient, with clear instructions and explanation on the Review and Investigation phases. The full prompt and details of our evaluation task are outlined in Appdx \ref{app:evaluation_framework}.

\subsection{Results} 
\begin{table}[!tbp] 
    \vspace{-4\baselineskip}
    {
    \centering
    \footnotesize 
    \caption{Comparative performance of evaluated LLMs on VivaBench.}
    \label{tab:model_performance_metrics}
    \setlength{\tabcolsep}{-1pt} 

    \begin{tabularx}{\columnwidth}{@{} L{1.0in} *{3}{Y} *{3}{Y} *{3}{Y} *{3}{Y} *{2}{Y} *{2}{Y} @{}} 
    \toprule
    \multirow{3}{*}{\textbf{Model}} &
    \multicolumn{3}{c}{\textbf{Actions}} &
    \multicolumn{3}{c}{\textbf{Top-k P. ($d_P$)}} &
    \multicolumn{3}{c}{\textbf{Top-k D. ($d_D$)}} &
    \multicolumn{3}{c}{\textbf{Top-k F. ($d_F$)}} &
    \multicolumn{2}{c}{\textbf{Raw Conf.}} &
    \multicolumn{2}{c}{\textbf{$S_{\text{conf}}$}} \\
    \cmidrule(lr){2-4} \cmidrule(lr){5-7} \cmidrule(lr){8-10}
    \cmidrule(lr){11-13} \cmidrule(lr){14-15} \cmidrule(lr){16-17}
    & \textbf{Total} & \textbf{Rv} & \textbf{Ix}
    & \textbf{1} & \textbf{3} & \textbf{5}
    & \textbf{1} & \textbf{3} & \textbf{5}
    & \textbf{1} & \textbf{3} & \textbf{5}
    & \textbf{$C_P$} & \textbf{$C_D$}
    & \textbf{$S_P$} & \textbf{$S_D$} \\
    \midrule

    
        Gemini 2.5 Pro  &  8.8  &  4.4  &  2.3  &  0.17  &  0.30  &  0.33  &  0.35  &  0.46  &  0.48  &  0.69  &  0.84  &  0.86  &  0.45  &  0.77  &  0.25  &  0.37  \\
        DeepSeek-R1  &  5.5  &  2.0  &  1.5  &  0.12  &  0.22  &  0.24  &  0.23  &  0.31  &  0.32  &  0.61  &  0.77  &  0.80  &  0.33  &  0.59  &  0.22  &  0.35  \\
        o4-mini  &  8.9  &  4.0  &  2.9  &  0.15  &  0.24  &  0.26  &  0.32  &  0.40  &  0.41  &  0.63  &  0.79  &  0.81  &  0.38  &  0.78  &  0.25  &  0.44  \\
        Llama-4 Maverick  &  8.5  &  3.0  &  3.5  &  0.08  &  0.17  &  0.18  &  0.23  &  0.27  &  0.27  &  0.52  &  0.70  &  0.71  &  0.49  &  0.70  &  0.20  &  0.37  \\
        Grok 3 Mini Beta  &  7.0  &  3.4  &  1.6  &  0.10  &  0.20  &  0.23  &  0.16  &  0.27  &  0.29  &  0.60  &  0.76  &  0.77  &  0.30  &  0.36  &  0.20  &  0.26  \\
        Qwen 3  &  5.5  &  1.8  &  1.8  &  0.10  &  0.19  &  0.21  &  0.21  &  0.30  &  0.31  &  0.47  &  0.66  &  0.71  &  0.29  &  0.47  &  0.20  &  0.29  \\
    \bottomrule
    \end{tabularx}}
    \footnotesize 
    \textbf{Actions}: Average interaction turns per case (Total), broken down by Review queries ($Rv = \mathcal{H}+\mathcal{P}$) and Investigation ($Ix = \mathcal{I}+\mathcal{L}$)  orders.
    \textbf{Top-k}: Top-k Exact Accuracy for Provisional ($d_P$) \textbf{(P)}, Final ($d_D$) \textbf{(F)} (after interactive information gathering), and Full Information ($d_F$) \textbf{(F)} Diagnoses at $k \in \{1,3,5\}$.
    \textbf{$C_P, C_D$}: Mean raw confidence scores for provisional and final diagnoses, respectively.
    \textbf{$S_P, S_D$}: Confidence-Weighted Accuracy Scores ($S_{\text{conf}}$) for provisional and final diagnoses, respectively.
\end{table}

    \begin{figure}
    \centering
    \includegraphics[width=0.5\linewidth]{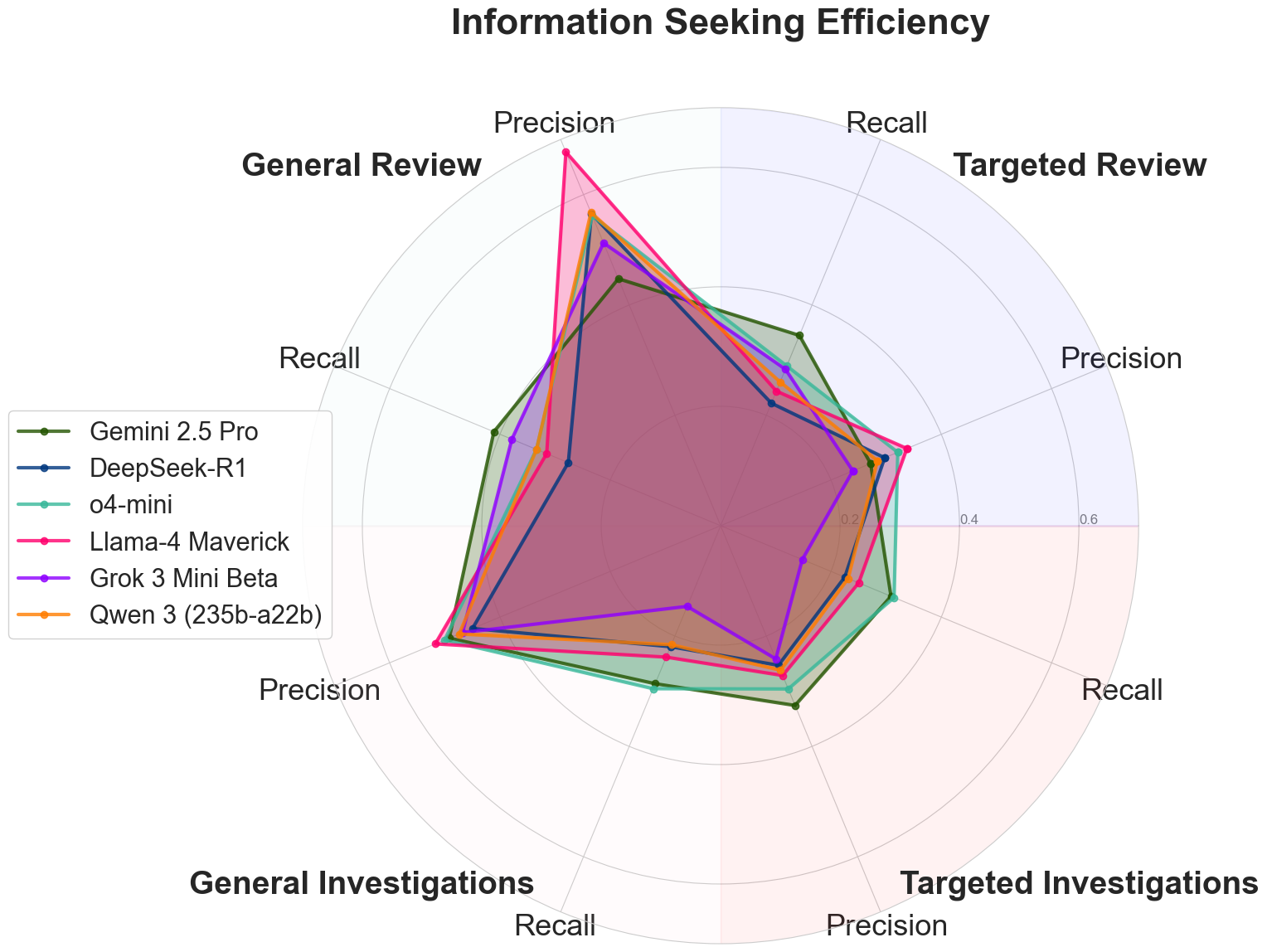}
    \caption{Radar plot comparing precision and recall metrics for how effectively models gather clinical information. \textbf{Targeted} metrics assess performance on gathering diagnosis-relevant information only, while \textbf{Overall} metrics include all available clinical information. \textbf{Review} includes history-taking and physical examination, while \textbf{Investigations} covers labs and imaging. Higher values are better.}
    \label{fig:radar-plot}
    \vspace{-\baselineskip}
\end{figure}

\begin{figure}[!t]
\vspace{-2\baselineskip}
    \centering    \begin{minipage}[t]{0.54\textwidth}
        \centering
        \includegraphics[width=\linewidth]{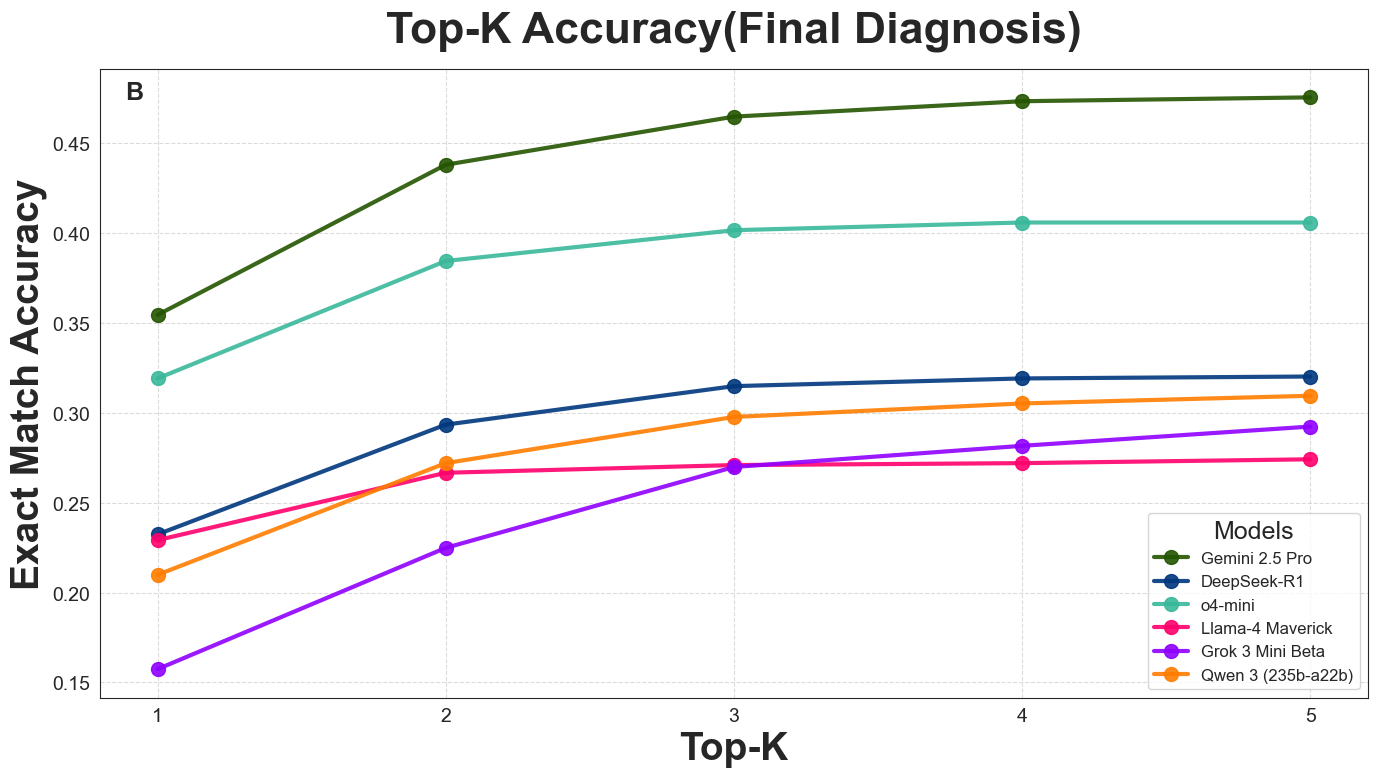}
        \caption{Accuracy of top-1 to top-5 diagnoses.}
        \label{fig:top-5-acc}
    \end{minipage}
    \hfill
    \begin{minipage}[t]{0.43\textwidth}
        \centering
        \includegraphics[width=\linewidth]{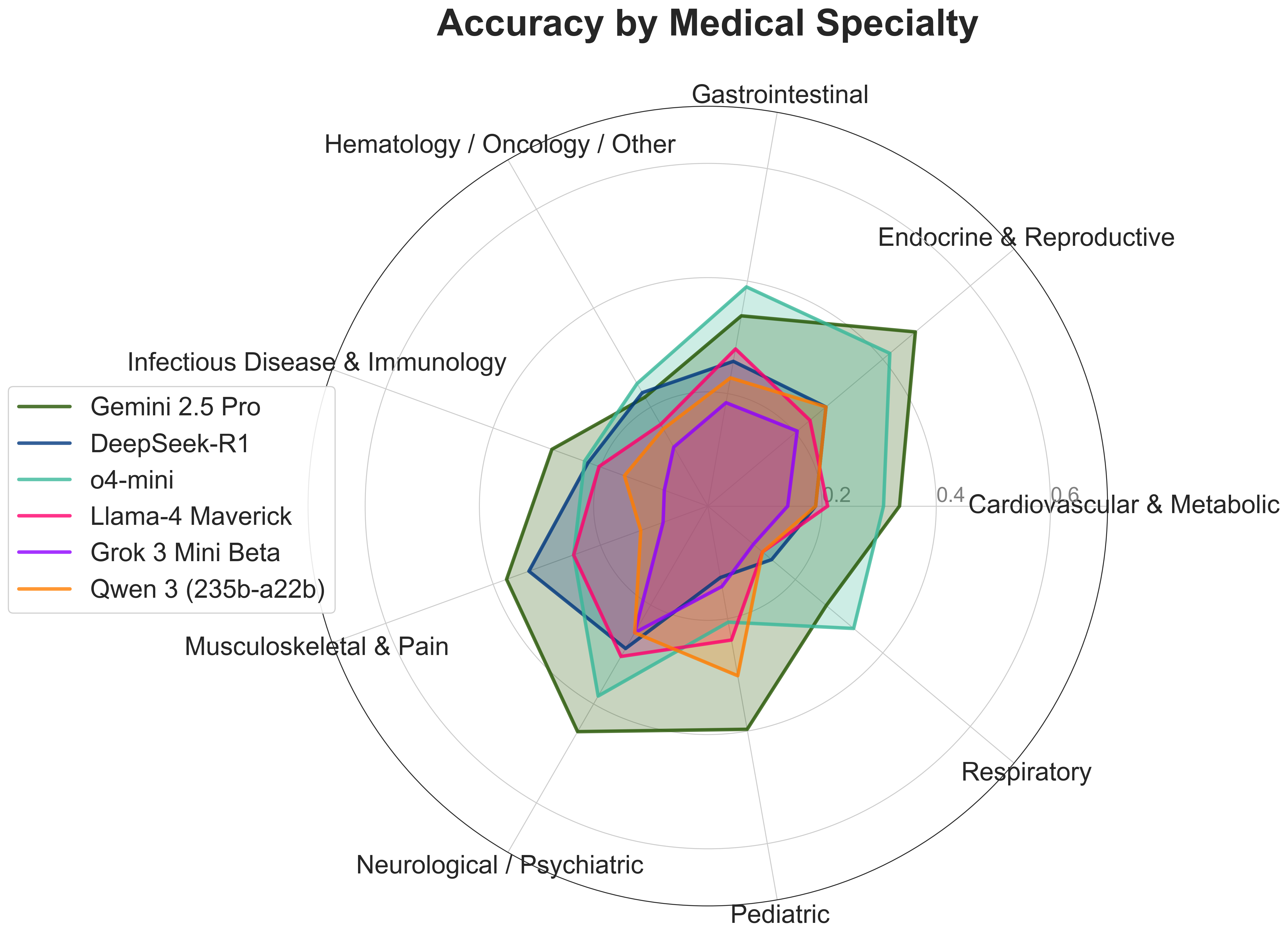}
        \caption{Performance across specialties.}
        \label{fig:specialty_acc}
    \end{minipage}
    \vspace{-\baselineskip}
\end{figure}

\textbf{Diagnostic Accuracy} The diagnostic performance results are outlined in \cref{tab:model_performance_metrics}. Overall, while most models demonstrated competence in deriving diagnoses when presented with complete clinical information (Full Information condition), they struggled significantly when constrained to the interactive examination format. Gemini 2.5 Pro consistently outperformed other models across all diagnostic stages and metrics, achieving the highest top-1 accuracy in final diagnosis (35\%) and full information scenarios (69\%). This performance difference can be attributed with the relative recency and scale of the models evaluated, suggesting that larger, more recent models may better handle the complex reasoning required for sequential diagnostic tasks. Nonetheless, there is a substantial performance gap between final diagnosis accuracy and full information accuracy across all models, with most models at least doubling in accuracy performance. This suggests that models possess the requisite knowledge to come to a diagnosis (or straightup has the data in the training set), but they struggle with handling uncertainty or to undergo the targeted information-seeking process needed to apply it effectively. The progression from top-1 to top-5 accuracy (Figure \ref{fig:top-5-acc}) shows diminishing returns across all models, with most significant gains occurring between top-1 and top-3. This suggests that when models are incorrect in their primary diagnosis, the correct answer is often present within their top alternatives, indicating partial understanding of the clinical picture.

Subgroup analysis demonstrated variation in model performance across different specialty groups (Figure \ref{fig:specialty_acc}). All models demonstrated relative strengths in Infectious Disease \& Immunology and Cardiovascular \& Metabolic conditions, likely reflecting the prevalence of these conditions in medical literature and their often distinctive presentation patterns. Conversely, most models showed comparative weakness in Pediatric and Neurological/Psychiatric conditions, which frequently require more nuanced interpretation of symptoms and developmental context.

\begin{table}[!htbp]
    \centering
    \footnotesize 
    \caption{\small Correlation analysis of diagnostic adaptation behaviors. The table shows Pearson correlation coefficients between diagnostic adaptation \textbf{Variables} (center-right columns) and performance \textbf{Outcomes} (left data columns).
    \textbf{Outcomes} are defined as the change from provisional ($d_p$) to final ($d_f$) stage in: Top-1 Exact Accuracy (Acc $\Delta$), Top-1 Approximate Accuracy (Apr. Acc $\Delta$), and Confidence-Weighted Score ($S_{\text{conf}}$ $\Delta$).
    \textbf{Variables} include: counts of Diagnoses Added, Removed, or Maintained; Confidence Delta (overall change in mean raw confidence, $C_D - C_P$); Confidence Shift (mean $c_f - c_p$) and Confidence Shift Magnitude (mean $|c_f - c_p|$) for maintained diagnoses.
    Superscripts indicate statistically significant correlations ($p < 0.05$): \textsuperscript{\textdaggerdbl} with Exact Accuracy Change; \textsuperscript{\textdagger} with Approximate Accuracy Change; \textsuperscript{*} with $S_{\text{conf}}$ Change.\\[-0.5\baselineskip]}
    \label{tab:correlation_analysis} 
    \setlength{\tabcolsep}{3pt} 
    
    \noindent 
    \makebox[\columnwidth][c]{%
    \begin{tabularx}{1.2\columnwidth}{@{} L{1.1in} *{9}{y} @{}}
    \toprule
    \multirow{2}{*}{\stackheader{Model}} &
    \multicolumn{3}{c}{\stackheader{Outcomes (Change $d_p \rightarrow d_f$)}} &
    \multicolumn{6}{c}{\stackheader{Diagnostic Adaptation Variables}} \\
    \cmidrule(lr){2-4} \cmidrule(lr){5-10} 
    & \stackheader{Acc\\$\Delta$} & \stackheader{Apr.\\Acc $\Delta$} & \stackheader{$S_{\text{conf}}$\\$\Delta$} & \stackheader{Diag.\\Added} & \stackheader{Diag.\\Removed} & \stackheader{Diag.\\Maintained} & \stackheader{Conf.\\Delta} & \stackheader{Conf.\\Shift} & \stackheader{Conf.\\Shift Mag.} \\
    \midrule

    Gemini 2.5 Pro    & 0.18 & 0.09 & 0.12 & $2.25^{*}$ & $3.08^{\dagger\ddagger*}$ & $0.76^{\dagger\ddagger*}$ & $0.32^{\dagger\ddagger*}$ & 0.10 & $0.19^{*}$ \\
    \addlinespace
    DeepSeek-R1       & 0.11 & 0.08 & 0.13 & $1.69^{\dagger\ddagger*}$ & $2.65^{\dagger\ddagger*}$ & $0.90^{\dagger\ddagger*}$ & $0.26^{\dagger\ddagger*}$ & $0.12^{\dagger\ddagger*}$ & 0.24 \\
    \addlinespace
    o4-mini           & 0.17 & 0.10 & 0.18 & $1.24^{\ddagger}$ & $2.47^{\dagger\ddagger*}$ & $0.70^{\dagger\ddagger*}$ & $0.40^{\dagger\ddagger*}$ & 0.12 & $0.20^{*}$ \\
    \addlinespace
    Llama-4 Maverick  & 0.14 & 0.10 & 0.17 & 1.11 & $1.79^{\dagger\ddagger*}$ & $0.71^{\dagger\ddagger*}$ & $0.21^{\dagger\ddagger*}$ & $0.05^{\dagger\ddagger*}$ & 0.17 \\
    \addlinespace
    Grok 3 Mini Beta  & 0.05 & 0.03 & 0.06 & $1.40^{\ddagger}$ & $1.64^{\dagger\ddagger*}$ & $2.70^{\dagger\ddagger*}$ & $0.05^{\dagger\ddagger*}$ & $0.00^{\ddagger}$ & $0.42^{\ddagger}$ \\
    \addlinespace
    Qwen 3 (235b-a22b)& 0.11 & 0.06 & 0.08 & $1.94^{\dagger\ddagger}$ & $2.55^{\dagger\ddagger*}$ & $1.34^{*}$ & $0.18^{\dagger\ddagger*}$ & $0.07^{\ddagger}$ & $0.27^{*}$ \\
    \bottomrule
    \end{tabularx}
    }
    \vspace{-\baselineskip}
\end{table}

\textbf{Information Seeking Efficiency}
To evaluate how models gather clinical information strategically, we measured their precision and recall across different information-seeking activities. We define "information seeking efficiency" as the ability to selectively request relevant history, physical examination findings, and diagnostic investigations that contribute to accurate diagnosis. This provides insight into the diagnostic strategies models employ—precision reflects how targeted their inquiries are toward specific diagnoses, while recall indicates their thoroughness in confirming diagnoses and excluding alternatives.

Figure \ref{fig:radar-plot} shows distinct information-seeking patterns across models. Gemini 2.5 Pro demonstrated the most balanced approach, with strong performance in both overall information gathering (breadth) and diagnosis-targeted inquiries (depth). This balanced information-seeking strategy likely contributed to its superior diagnostic accuracy. Most models showed asymmetric patterns, with generally higher precision than recall across all categories, suggesting they are selective in their inquiries but often miss relevant clinical details.

Models generally performed better at selecting appropriate clinical review items than ordering relevant investigations. The low recall in diagnostic investigations could be attributed to the abundance of clinical information within the vignettes. However,  their precision were also low in terms of ordering the exact tests that would help them with diagnoses. While a lot of the requested tests were basic routine blood tests that were not included in the vignette, we also observe that models were ordering a lot of random and super niche stuff, such as genetic testing. It is likely that this was due to training, where models would naturally be trained on the more specific / niche stuff in an academic setting. 

\textbf{Diagnostic Adaptation} To further quantify how models adjust their decision making process in light of uncertainty, we measured their diagnostic adaptation behaviors and performance outcomes. In particular, we measured changes in diagnoses between provisional and final stages, and also how the model's confidence distribution changed between steps. We tracked absolute confidence changes, defined as the difference averaged value of confidence of diagnoses between stages, and the shift of confidence in maintained diagnoses (diagnoses sustained between provisional and final diagnosis stage). Table \ref{tab:correlation_analysis} shows correlations between diagnostic adaptation behaviors and performance outcomes, as measured by the delta change in exact and top-1 accuracy, as well as confidence S.

\begin{figure}[t]        \centering\small
        \includegraphics[width=0.42\linewidth]{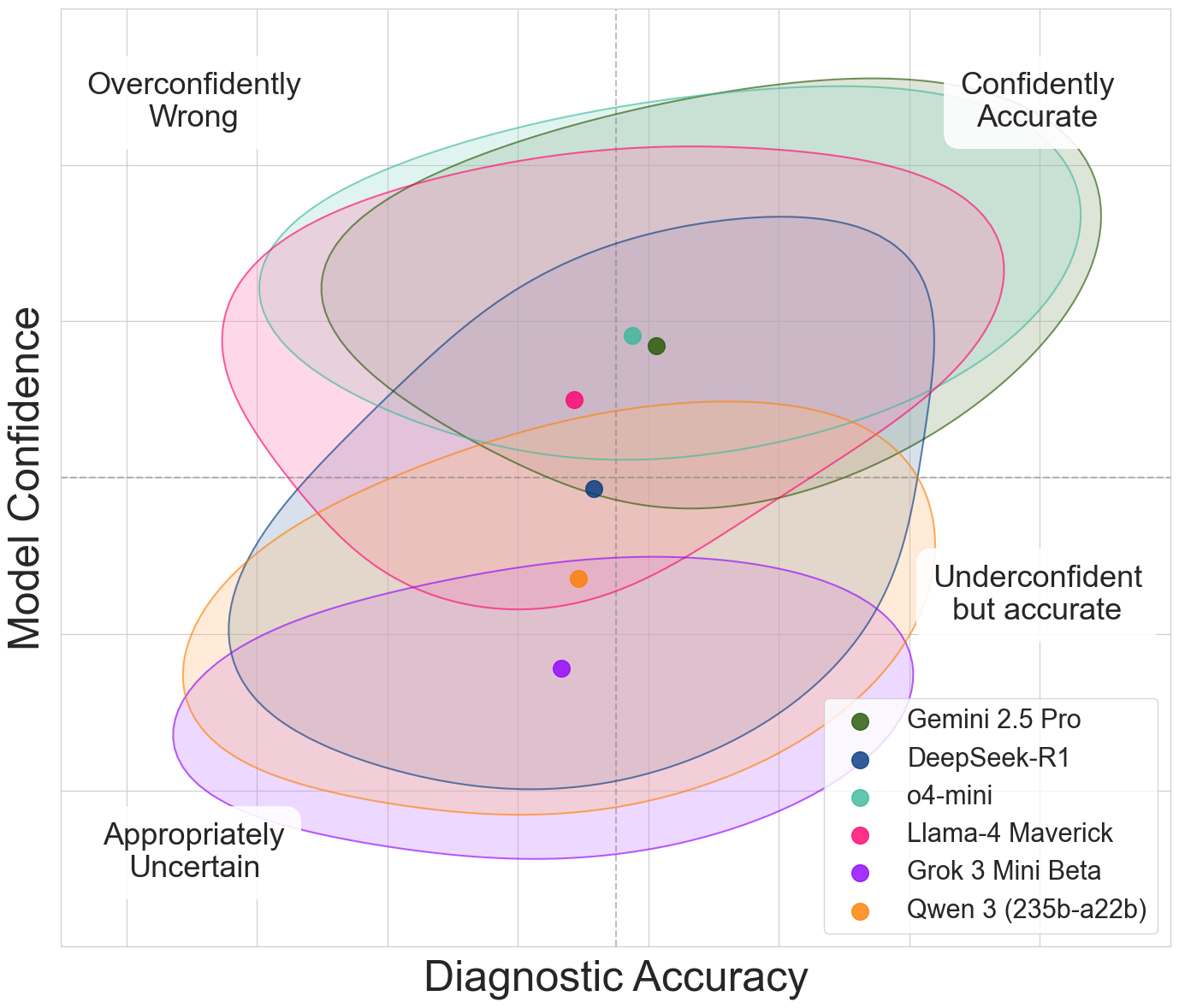}
        \caption{Kernel density estimation (KDE) contours representing the distribution model performance across clinical cases. The quadrants reflect differences in confidence and accuracy of diagnosis: confidently accurate (top-right), overconfident and wrong (top-left), underconfident but accurate (bottom-right), and appropriately uncertain (bottom-left).\\[-3\baselineskip]}
        \label{fig:confidence-accuracy-matrix}
\end{figure}

Diagnosis removal and diagnosis maintenance both showed significant correlation to all 3 outcome metrics. This is expected, as models that correctly identified diagnoses during the provisional stage only needed to further investigate and rule out their other suspected conditions. In contrast, diagnosis addition showed less significant correlation to outcomes, representing a more challenging adaptation task. We observed divergent patterns between model performance tiers: lower-performing models showed improved accuracy but not confidence when adding diagnoses, suggesting more random but less confident guesses. Better-performing models showed improvements in both metrics or primarily in confidence scores, indicating they could recognize when initial diagnostic considerations were incomplete and adjust accordingly.

\textbf{Confidence} A consistent pattern across all models was the strong positive correlation between absolute confidence changes and diagnostic performance improvements. As models gathered more information, those that demonstrated larger shifts in overall confidence (regardless of direction) tended to show greater improvements in both exact and approximate diagnostic accuracy. This suggests that meaningful engagement with new clinical information is reflected in confidence recalibration and serves as an indicator of effective diagnostic reasoning.

All models tended to increase confidence in diagnoses they maintained between stages, regardless of whether those diagnoses were correct. The magnitude of confidence change for maintained diagnoses correlated strongly with overall confidence score changes, but showed limited correlation with accuracy improvements. This suggests that models' growing certainty often developed independently of diagnostic correctness, indicating a form of confirmation bias.

DeepSeek and Llama displayed significant correlations between the direction of confidence shifts and accuracy changes, suggesting more calibrated confidence adjustments, where they increase confidence primarily when moving toward correct diagnoses. This behavior appears related to their information-seeking efficiency, as both models demonstrated higher precision in their review and diagnostic metrics, suggesting that their ability to seek targeted information allows them to hone in on the correct diagnosis, while increasing their confidence in doing so.

To further explore the relationship between diagnostic performance and confidence calibration, we mapped models' confidence against accuracy using KDE distributions (Figure \ref{fig:confidence-accuracy-matrix}). This analysis revealed distinct calibration across model architectures. Gemini 2.5 Pro and o4-mini demonstrated well-calibrated confidence, positioning primarily in the "confidently accurate" quadrant. DeepSeek-R1 exhibited more conservative confidence estimation despite reasonable accuracy, while Llama-4 Maverick showed a more complex pattern with higher overall confidence despite mixed accuracy results. Grok 3 Mini Beta and Qwen 3 displayed lower confidence profiles, and had a diverse range of accuracy metrics across the full dataset. These distinctions highlight fundamental differences in how models internally calibrate uncertainty during diagnostic reasoning, with important implications for clinical deployment where both accurate diagnoses and appropriate confidence signaling are essential.

\subsection{Failure Mode Analysis}
To explore why agents failed to perform clinical reasoning over, we first qualitatively reviewed the interation / reasoning trace between agents, and we came up with three common failure modes

\textbf{Quantitative analysis} To further explore the frequency, we used an LLM as classifier to group X cases into groups, and if there were any other major reasons. 
The finding breakdown includes inappropriate hypothesis generation (348 cases), premature diagnostic closure (291), inadequate investigations (90), and ICD coding errors (9). These findings reveal that the majority of model failures are attributed to reasoning issues reflecting common failure modes in clinical practice (anchoring bias, satisfaction-of-search bias), while limitations due to knowledge (e.g. ordering appropriate investigations) being less of an issue. 
 
\textbf{Qualitative Analysis} To supplement our findings, we reviewed individual agent reasoning trajectories to identify patterns in clinical decision-making failures. We present several cases in Appendix \ref{app:qualitative} that illustrate the failure modes we observed throughout our evaluations. Collectively, these examples highlight potential scenarios that could compromise patient safety through missed diagnoses, delayed interventions, or inappropriate clinical pathways if current AI agents are deployed in real-world settings.

\section{Discussion} 
Our evaluation reveals a significant gap between LLMs' knowledge base and their ability to engage in sequential clinical reasoning. The performance disparity between full-information and interactive examination conditions demonstrates that current models possess relevant medical knowledge, but struggle with systematic information gathering and hypothesis refinement under uncertainty. This leads to three patterns of failure modes, with inefficient information seeking (high precision but low recall), suboptimal hypothesis revision (difficulty adding appropriate new diagnoses), and confidence miscalibration (increasing confidence regardless of diagnostic correctness). These findings highlight how current LLMs, despite superior knowledge retrieval capabilities, exhibit reasoning patterns that diverge from sound clinical judgment when navigating diagnostic uncertainty. VivaBench connects to several active research areas including sequential decision-making, reasoning agents, confidence calibration, and strategic information gathering, and provides a structured environment for the wider machine learning community to study how models make decisions with partial information. 

\textbf{Limitations} Our study has several limitations. The dataset derives from a single source type (clinical case reports), and remains modest in scale compared to large-scale medical benchmarks, due to the labour intensivity of the data generation process. The determinstic variant of our information retrieval system does not capture the full variability in real clinical communication, while our LLM variant is not fully determinstic, affecting the validity of our benchmark. Additionally, due to computational constraints, we conducted only a single evaluation run for each model, which may not account for the stochastic nature of LLM outputs, especially over long horizons. Lastly, despite structuring cases according to clinical standards, the \textit{viva voce} examination format remains a simplified approximation of actual patient encounters, lacking the multi-faceted nature and nuance in clinical practice.

\newpage

{\small
\bibliography{refs}
}


\include{_appendix}



\end{document}

%% file: commands.tex
\ifcomments
    \newcommand{\TODO}[1]{{\color{red}{[TODO: #1]}}}
    \newcommand{\CC}[1]{{\color{blue}{[CC: #1]}}}
    \newcommand{\SP}[1]{{\color{teal}{[SP: #1]}}}

\else
    \newcommand{\TODO}[1]{}
    \newcommand{\CC}[1]{}
    \newcommand{\SP}[1]{}

\fi

%% file: _appendix.tex
\newpage
\appendix
\section{Schema Design}
\label{app:schema_design}

The base unit for each test case in the VivaBench benchmark is a Clinical Case ($\mathcal{C}$). Each Case is converted from a free-text clinical vignette leading up to a diagnosis, and comprises five primary components: History ($\mathcal{H}$), Physical Examination ($\mathcal{P}$), Imaging ($\mathcal{I}$), Laboratory investigations ($\mathcal{L}$), and a ground truth Diagnosis set ($\mathcal{D}$), which includes a list of accepted differential diagnoses ($\mathcal{D'}$).

\begin{algorithm}
\caption{\textbf{Input schema for a structured Clinical Case} $\mathcal{C}$}
\label{alg:schema_definition} 
\textbf{History} ($\mathcal{H}$):
\begin{itemize}[label={}, leftmargin=0.5em]
  \setlength\itemsep{-0.2em}
  \item \textbf{Demographics}: Patient age, gender, ethnicity.
  \item \textbf{Chief Complaint}: A one-line summary of the patient's presentation.
  \item \textbf{Symptoms}: A list of symptom objects, each with attributes including (but not limited to):
  \begin{itemize}
    \setlength\itemsep{-0.2em}
    \item \textbf{name}: Best matching SNOMED-CT description of the symptom.
    \item \textbf{onset}: When the symptom first began (e.g., "2 days ago").
    \item \textbf{duration}: How long the symptom has persisted (e.g., "3 hours").
    \item \textbf{progression}: How the symptom evolved over time.
    \item \textbf{timing}: When the symptom occurs (e.g., "morning", "after meals").
    \item \textbf{severity}: Intensity of the symptom (e.g., "mild", "severe").
    \item \textbf{system}: Body system affected (e.g., "cardiovascular").
    \item \textbf{location}: Anatomical location of the symptom.
    \item \textbf{character}: Quality or nature of the symptom (e.g., "sharp", "dull").
    \item \textbf{radiation}: Whether and where the symptom spreads.
    \item \textbf{alleviating\_factors}: Factors that improve the symptom.
    \item \textbf{aggravating\_factors}: Factors that worsen the symptom.
    \item \textbf{associated\_symptoms}: Other co-occurring symptoms.
    \item \textbf{context}: Circumstances surrounding the symptom.
    \item \textbf{history}: Detailed narrative specific to the symptom, extracted from the input vignette.
  \end{itemize}
  \item \textbf{Past Medical History}: Previous diagnoses and conditions.
  \item \textbf{Medication History}: Current medications with dosage and frequency.
  \item \textbf{Allergies}: Known drug or food allergies.
  \item \textbf{Social History}: Lifestyle factors (e.g., smoking, alcohol, occupation).
  \item \textbf{Family History}: Relevant conditions in family members.
\end{itemize}

\textbf{Physical Examination} ($\mathcal{P}$): Vital signs and physical findings, organized by body systems.

\textbf{Lab Investigations} ($\mathcal{L}$): Laboratory and diagnostic test results, typically mapped to LOINC codes.

\textbf{Imaging} ($\mathcal{I}$): Radiological studies, categorized by modality and anatomical region.

\textbf{Diagnosis} ($\mathcal{D}$): A list of final diagnosis objects. Each object includes:
\begin{itemize}
    \setlength\itemsep{-0.2em}
    \item The free-text label from the original vignette (e.g., "Tumefactive Crohn's disease").
    \item The best-matching ICD-10 code and name (e.g., K50.8, "Crohn's disease of both small and large intestine").
    \item A set of keys from $\mathcal{H}$, $\mathcal{P}$, $\mathcal{I}$, $\mathcal{L}$ that support this diagnosis.
\end{itemize}

\textbf{Differentials} ($\mathcal{D'}$): A list of alternative diagnosis objects. Each object includes:
\begin{itemize}
    \setlength\itemsep{-0.2em}
    \item Its own ICD-10 code and name (if applicable).
    \item A free-text field describing the clinical reasoning supporting it as a differential.
    \item (Optionally, supporting keys as in $\mathcal{D}$). 
\end{itemize}
\end{algorithm}

\noindent\textbf{Design considerations of clinical Schema.} To convert free-text clinical vignettes into a machine-readable format suitable for VivaBench, we defined the five-part schema for a structured clinical case $\mathcal{C} = \{\mathcal{H}, \mathcal{P}, \mathcal{I}, \mathcal{L}, \mathcal{D}\}$. $\mathcal{H}$ (History) captures the patient's clinical presentation and other relevant medical history. $\mathcal{P}$ (Physical Examination) records vital signs and examination findings, organized by body systems. $\mathcal{I}$ (Imaging) holds radiological and nuclear medicine studies, while $\mathcal{L}$ (Laboratory) contains all non-imaging investigations. $\mathcal{D}$ lists the final diagnosis (or diagnoses), and $\mathcal{D'}$ contains other accepted differential diagnoses. We refined this schema by reviewing standard clinical documentation guidelines \cite{RACGPClinicalGuidelines, ClinicalPracticeGuideline} and soliciting feedback from our clinician panel. Every field within the schema may be null or multi-valued, and corresponding excerpts from the input vignette are preserved for traceability to ensure fidelity to the source material.

\noindent\textbf{Overview of Clinical Schema.} Algorithm \ref{alg:schema_definition} describes our schema for $\mathcal{C}$ in detail.
Within $\mathcal{H}$, the presenting complaint is expressed as a list of Symptom objects. Each symptom is characterized by multiple attributes, with core attributes based on the SOCRATES framework \cite{gregory2019use} (e.g., Site, Onset, Character, Radiation, Associated symptoms, Time course, Exacerbating/relieving factors, Severity), supplemented by additional descriptive fields for more complex symptoms. When applicable, a time-stamped trajectory is included for any element that changes over time (e.g., "pain severity: mild $\rightarrow$ moderate $\rightarrow$ severe over 24 hours"). Relevant negative findings (e.g., "denies dyspnea") are also explicitly encoded to provide clinical context. $\mathcal{H}$ also includes distinct sub-sections for past medical and surgical history, current medications, allergies, social history, and family history.

All clinical concepts in $\mathcal{H}$ and $\mathcal{P}$ utilize SNOMED-CT codes \cite{SNOMEDCT} when available to ensure standardized terminology. Similarly, $\mathcal{L}$ and $\mathcal{I}$ items are mapped to LOINC procedural codes \cite{LOINC}. Final diagnoses ($\mathcal{D}$) and differentials ($\mathcal{D'}$) carry ICD-10 names and codes\cite{ICD10Version2019}. If a standard code is not available or applicable for any item, the original free-text label from the source vignette is used.

$\mathcal{D}$ can represent multiple final diagnoses if present in the source case. Each diagnosis entry specifies a condition name from the original vignette, its corresponding ICD-10 code and name, and the set of relevant keys from $\mathcal{H}$, $\mathcal{P}$, $\mathcal{I}$, and $\mathcal{L}$ that contribute to the diagnosis. A list of accepted differentials ($\mathcal{D'}$) is included for cases with inherent diagnostic ambiguity (e.g., when the original vignette considered multiple diagnoses) or those requiring specialized investigations (e.g., differentiating subtypes of lung cancer). In addition to their own ICD-10 mappings, each entry in $\mathcal{D'}$ includes a free-text field describing the clinical reasoning that supports it as a differential diagnosis.

Our schema aggregates data across these domains from a cross-sectional, first-visit perspective. Information regarding prior admissions or procedures not directly relevant to the patient's current presentation is captured under past medical history within $\mathcal{H}$. Repeat admissions for the same presenting complaint are merged into the initial encounter by appending trajectories to the relevant data fields. For example, a patient who had open heart surgery two months prior and was discharged yesterday for stable angina, then readmitted with recurrent chest pain hours later, will have "open heart surgery 2 months ago" in the past medical history section of $\mathcal{H}$. All clinical information directly related to the current presentation (including the recurrent chest pain) is aggregated into their respective single fields within the current encounter, with temporal evolution preserved (e.g., symptoms: "Chest pain initially improved with GTN. However, re-emerged 24 hours later"; investigations: "Troponin: 5 $\rightarrow$ 7 $\rightarrow$ 11 ng/mL at 1h, 3h, 5h").
\section{Dataset Generation Pipeline}
\label{app:dataset_generation}

\begin{figure}[htb] 
    \centering
    \includegraphics[width=1\linewidth]{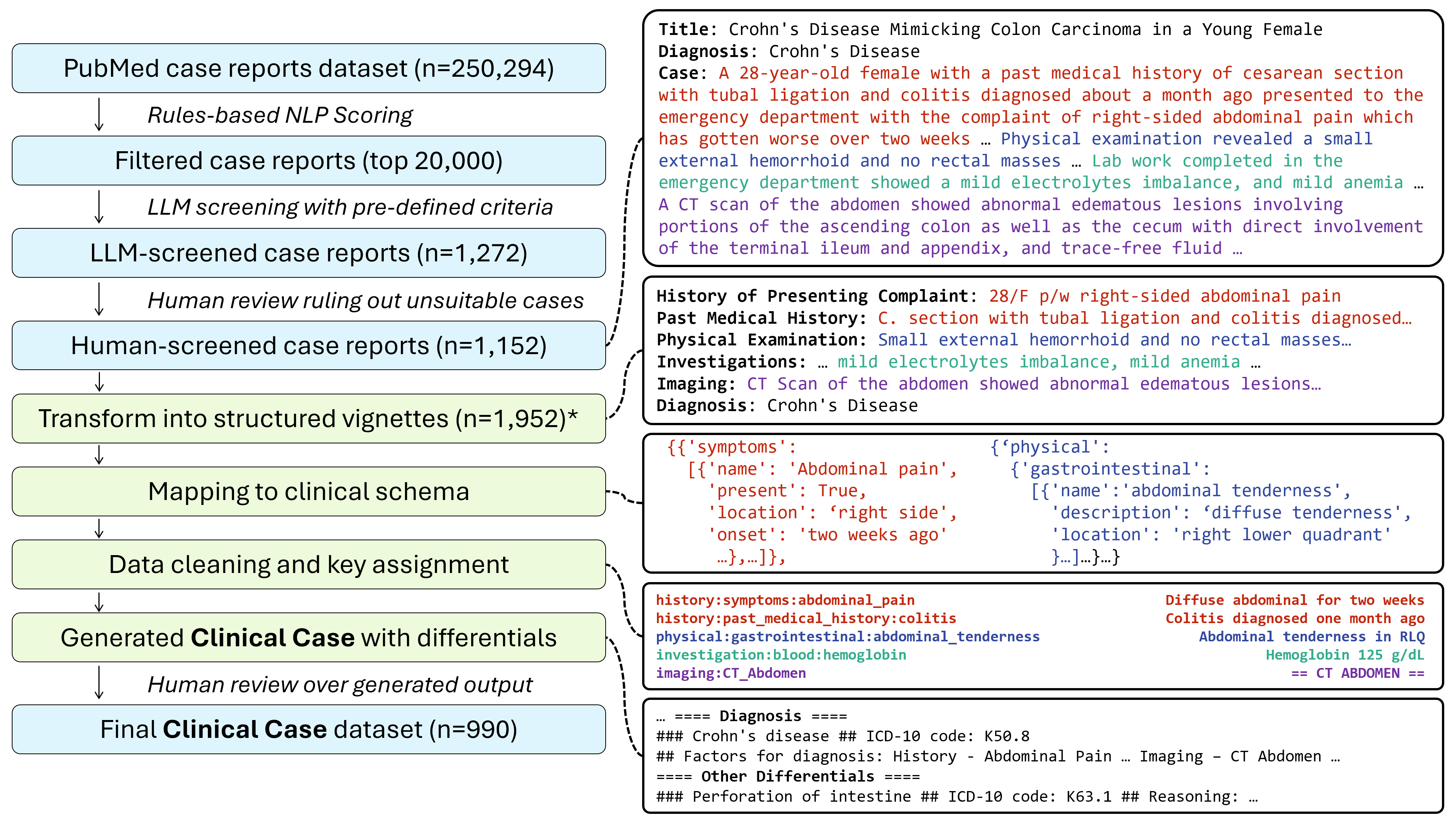}
    \caption{Our clinical case dataset generation pipeline, transforming PubMed case reports into structured, machine-learning-ready clinical vignettes for VivaBench. The workflow (left panel) illustrates filtering stages (blue boxes) that reduce an initial pool of 250,294 raw PubMed reports to 1,152 human-screened candidate cases, and subsequent transformation steps (green boxes) that result in 990 final structured clinical cases (*Additional cases from MedQA dataset were also generated in our initial experiments). The right panel provides an example of this transformation for a 28-year-old female with Crohn's Disease, showing the conversion of unstructured medical text into a standardized case format. This structured representation preserves the clinical reasoning process essential for our evaluation framework.}
    \label{fig:dataset-generation-pipeline}
\end{figure}

We employed a hybrid approach combining automated Natural Language Processing (NLP) heuristics, LLM-based screening, and human expert review to generate the VivaBench dataset. Figure \ref{fig:dataset-generation-pipeline} summarizes our dataset generation pipeline. The generation process begins with rules-based scoring over clinical case report databases to cases with comprehensive clinical information. The top scoring cases subsequently go through LLM-based screening against a set of pre-defined criteria to identify potentially relevant candidates. Human medical experts further review these candidates to ensure clinical accuracy and relevance. Selected clinical vignettes undergo transformation into our standardized schema detailed outlined in Appendix \ref{app:schema_design}. Lastly, generated clinical cases undergo another round of human review before inclusion to the final dataset.

\textbf{Dataset Source} To source clinical vignettes for VivaBench, we prioritized publicly accessible repositories providing free-text clinical data that included patient history, physical examination findings, investigation results, and clear diagnosis. Our exploration covered MedQA \cite{jin2021disease}, training materials from physician colleges in Australia and the United Kingdom \cite{SampleScenariosFederationa}, and case reports from PubMed \cite{2022NcbiPubmedDatasets}. We also examined restricted-license patient datasets, such as MIMIC \cite{johnson2016mimic}, during initial experiments to inform our schema design and data exploration strategies. Following a comprehensive review by our clinical team, PubMed case reports were selected as the primary source due to their wide variety, high quality of clinical detail, and rich diagnostic narratives spanning diverse medical specialties.

\textbf{Screening and Filtering} We implemented a two-stage filtering pipeline to identify the most educationally valuable cases from an initial pool of over 250,000 PubMed case reports. The first stage involved an initial screening pass using NLP techniques to analyze case report titles and content. This step aimed to identify reports containing substantive clinical review components and a clear diagnostic focus, while filtering out non-clinical reports, non-human subject studies, reports primarily describing surgical techniques, or those detailing exceptionally rare diseases. This rule-based filtering yielded approximately 20,000 candidate cases.

These candidate cases then underwent a second screening stage using an LLM. The LLM evaluated each case against predefined criteria, including: (1) richness of clinical information (presence of history, physical examination, and laboratory findings); (2) clarity of diagnostic reasoning presented; (3) appropriateness for general medical practice settings; (4) presence of diagnostic challenges or documentation of missed diagnoses; and (5) overall educational value. This LLM-based screening prioritized common conditions presented as educational resources, atypical presentations of common conditions, and diagnostically challenging scenarios, while excluding cases focused primarily on treatment outcomes, rare genetic disorders, or highly specialized testing. We filtered for cases that match at least four out of the five criteria, and 1,272 candidate cases. From these, 1,152 vignettes were selected after an initial human review for further processing. To balance between different specialty groups, we also included cases from the MedQA dataset \cite{jin2021disease} in our dataset generation pipeline (n = 1,952).

\textbf{Dataset Generation} Each case that passed initial screening underwent a NLP-based transformation process to convert free-text clinical vignette into the structured sections, including patient demographics, presenting complaints, medical history, physical examination findings, laboratory results, imaging findings, and diagnostic conclusions. Each section are mapped to our schema using a LLM-based system, which was prompted with section-specific instructions and few-shot examples to identify, categorize, and map unstructured clinical information from the narrative text to the corresponding fields in our structured schema. For our screening and filtering process, we used \texttt{gpt-4.1} with \texttt{temperature = 0}. 

\textbf{Clinical Validation} The VivaBench dataset underwent clinical validation at multiple stages to ensure high fidelity and accuracy. Our volunteer review panel consisted of six clinicians, including one primary care physician, one emergency physician, one intensive care specialist, one anaesthetist, one ophthalmologist, and an internal medicine physician. Three medical students were also enrolled to assist with data parsing under clinical supervision. All volunteer reviewers will be acknowledged in the final camera ready.

During the initial case screening, clinicians assessed each candidate case for clinical accuracy, appropriate difficulty level for evaluating reasoning, and educational value. After conversion to structured format, clinicians reviewed the parsed information for accuracy, completeness, and appropriate categorization of all clinical elements. Any inaccuracies, inconsistencies, and missing information were manually corrected before finalization of dataset. The list of diagnoses and accepted differentials were also manually reviewed and annotated.

\begin{table}
    \centering
    \begin{tabular}{ccc}
         \textbf{Journal} &  \textbf{Specialty} & \textbf{Count}\\
         \hline
         Cureus &  General & 226\\
         J Med Case Rep &  General & 64\\
         Eur Heart J Case Rep &  Cardiology & 31\\
         Medicine (Baltimore) &  General & 25\\
         Case Rep Med &  General & 18\\
         Clin Case Rep &  General & 18\\
         Case Rep Rheumatol &  Rheumatology & 18\\
         J Investig Med High Impact Case Rep &  General & 17\\
         SAGE Open Med Case Rep &  General & 17\\
 BMC Infect Dis & Infectious Diseases &17\\
 J Community Hosp Intern Med Perspect & General &13\\
 Case Rep Infect Dis & Infectious Diseases &9\\
 Case Rep Neurol & Neurology &9\\
 J Int Med Res & Internal Medicine &8\\
 Case Rep Emerg Med & Emergency Medicine &8\\
 World J Clin Cases & General &8\\
 J Med Cases & General &8\\
 Front Neurol & Neurology &8\\
 Case Rep Cardiol & Cardiology &7\\
 Case Rep Endocrinol & Endocrinology &7\\
    \end{tabular}
    \caption{Top 20 journals contributing case reports to VivaBench. The final dataset is sourced from 260 journals across all medical specialties}
    \label{tab:journal_dist}
\end{table}

\section{Dataset Composition} 
The dataset was selected from an initial pool of 250,294 PubMed case reports, involved rule-based NLP filtering (yielding $\sim$20,000 cases), LLM-based screening (yielding 1,272 cases), and an initial human review (selecting 1,152 PubMed cases). These, along with 800 MedQA cases, were structured, creating a pool of 1,952 cases. After clinician validation, our evaluation dataset for VivaBench consists of 990 cases across nine specialty groups: Endocrine \& Reproductive (150), Infectious Disease \& Immunology (150), Cardiovascular \& Metabolic (148), Gastrointestinal (147), Neurological/Psychiatric (136), Hematology/Oncology/Other (112), Pediatric (69), Respiratory (51), and Musculoskeletal \& Pain (26). Both the complete set of 1,952 generated structured cases and the curated evaluation dataset are made available on our Hugging-Face repository: 
https://huggingface.co/datasets/chychiu/VivaBench/

The final dataset was sourced from 260 different journals, with Table \ref{tab:journal_dist} showing the top 20 sources by count, and their respective specialties. The journal distribution largely reflects the underlying prevalence in medical case report literature, with general medical journals predominating alongside specialty-specific publications. We note two outliers (Cureus, J Med Case Rep) from this dataset, both of which were aggregators of case reports or smaller medical publications. 

\textbf{Human Baseline} To assess the difficulty and appropriateness of dataset in practice, we obtained a brief human baseline, where 4 clinicians provided a provisional diagnosis based on the clinical picture over 14 unseen cases. Our clinician baseline achieved a 0.68 ± 0.09 top-3 accuracy versus 0.52 ± 0.07 for models at the provisional stage, suggesting our benchmark approximates clinician-level expectations. A larger scale human baseline would be one valuable direction for future research. 

\subsection{Analysis on Dataset Contamination}
\label{sec:dataset_contamination} As VivaBench was derived from existing published medical journal, there is concern dataset contamination affecting benchmark validity and results. To assess whether evaluated models had memorized complete case reports from the training data, we conducted a sentence completion experiment following the methodology of Golchin et al. \cite{golchinTimeTravelLLMs2024}.

\textbf{Methodology} We extracted the title and first sentence from each case report in our PubMed dataset and prompted models to generate the second sentence as it appeared in the original publication. We referencing the system prompt described in \cite{golchinTimeTravelLLMs2024} for our system prompt: \begin{quote}
\textit{``You are provided with the Title and Sentence 1 from a case report from \{dataset\}. Finish Sentence 2 as appeared in the dataset. Sentence 2 must exactly match the instance in the dataset.''}
\end{quote}

We evaluated two prompting strategies: (1) generic prompting with \texttt{dataset="PubMed database"}, and (2) journal-specific prompting using the actual source journal name. We used BLEURT~\cite{sellamBLEURTLearningRobust2020} and ROUGE-L \cite{linROUGEPackageAutomatic} to quantify textual similarity between generated and original sentences. In addition, we tested whether models could extract journal names from titles alone or author lists from journal and title combinations.

\textbf{Results} Generated sentences exhibited consistently low textual similarity to the original second sentences across all models and experimental configurations (Table~\ref{tab:completion_scores}). 
Manual review of the top 10\% of cases ranked by either similarity metric revealed no exact matches with the original publications. 
Furthermore, models failed to accurately extract journal names or author lists, and frequently hallucinated references during both reasoning and output generation.

\begin{table}
\centering
\begin{tabular}{l l l l l l}
model & setting & bleurt & ROUGE-F1 & ROUGE-prec & ROUGE-recall \\
llama-4-maverick & general & 0.25 +/- 0.08 & 0.12 +/- 0.08 & 0.11 +/- 0.09 & 0.15 +/- 0.13 \\
 & journal & 0.25 +/- 0.08 & 0.12 +/- 0.09 & 0.11 +/- 0.10 & 0.17 +/- 0.14 \\
grok-3-mini & general & 0.25 +/- 0.07 & 0.12 +/- 0.09 & 0.13 +/- 0.10 & 0.14 +/- 0.11 \\
 & journal & 0.25 +/- 0.06 & 0.12 +/- 0.08 & 0.13 +/- 0.09 & 0.12 +/- 0.08 \\
o4-mini & general & 0.22 +/- 0.08 & 0.10 +/- 0.09 & 0.11 +/- 0.10 & 0.12 +/- 0.11 \\
 & journal & 0.21 +/- 0.07 & 0.10 +/- 0.07 & 0.10 +/- 0.09 & 0.12 +/- 0.09 \\
deepseek-r1 & general & 0.28 +/- 0.06 & 0.08 +/- 0.06 & 0.06 +/- 0.06 & 0.20 +/- 0.12 \\
 & journal & 0.26 +/- 0.05 & 0.09 +/- 0.08 & 0.07 +/- 0.07 & 0.19 +/- 0.14 \\
qwen3-235b-a22b & general & 0.24 +/- 0.07 & 0.10 +/- 0.07 & 0.09 +/- 0.08 & 0.13 +/- 0.10 \\
 & journal & 0.24 +/- 0.06 & 0.10 +/- 0.08 & 0.10 +/- 0.09 & 0.13 +/- 0.09 \\
gemini-2.5-pro & general & 0.28 +/- 0.08 & 0.15 +/- 0.12 & 0.18 +/- 0.15 & 0.16 +/- 0.13 \\
 & journal & 0.28 +/- 0.08 & 0.18 +/- 0.13 & 0.21 +/- 0.16 & 0.18 +/- 0.14 \\

\end{tabular}
\caption{The BLEURT, ROUGE-F1, precision and recall for with generic prompt (\textit{general}) and provision of journal title (\textit{journal}) in the prompt}
\label{tab:completion_scores}
\end{table}

These findings suggest that while the case reports were published prior to the training data cutoff dates, the reports were unlikely to be directly memorised by 

\section{Information Retrieval and Parsing}
\label{app:ir_parsing}
A core technical challenge in our framework is the translation between natural language clinical queries generated by an agent and the structured information available in a clinical case $\mathcal{C}$. To address this, we developed a two-stage approach consisting of a Mapper module ($\mathbb{M}$) and a Parser module ($\mathbb{P}$).

The Mapper $\mathbb{M}$ functions as a translation mechanism. It maps any free-text clinical query $\mathcal{Q}$ originating from an agent's action in the space $\mathcal{A} = \{\mathcal{H}, \mathcal{P}, \mathcal{I}, \mathcal{L}\}$ (representing history, physical examination, imaging, and lab investigations respectively) to a set of structured information keys $\mathcal{K}_Q \subseteq \mathcal{K}_C$, where $\mathcal{K}_C$ denotes the complete set of information keys available within a clinical case $\mathcal{C}$. For each $k_i \in K_C$, there is a corresponding value $v_i \in \mathcal{V}_C$, which represents the set of all associated textual values for these keys. These key-value pairs follow the structure of our clinical schema, and defined through our data generation process.

\begin{figure}[htb]
    \centering
    \includegraphics[width=1\linewidth]{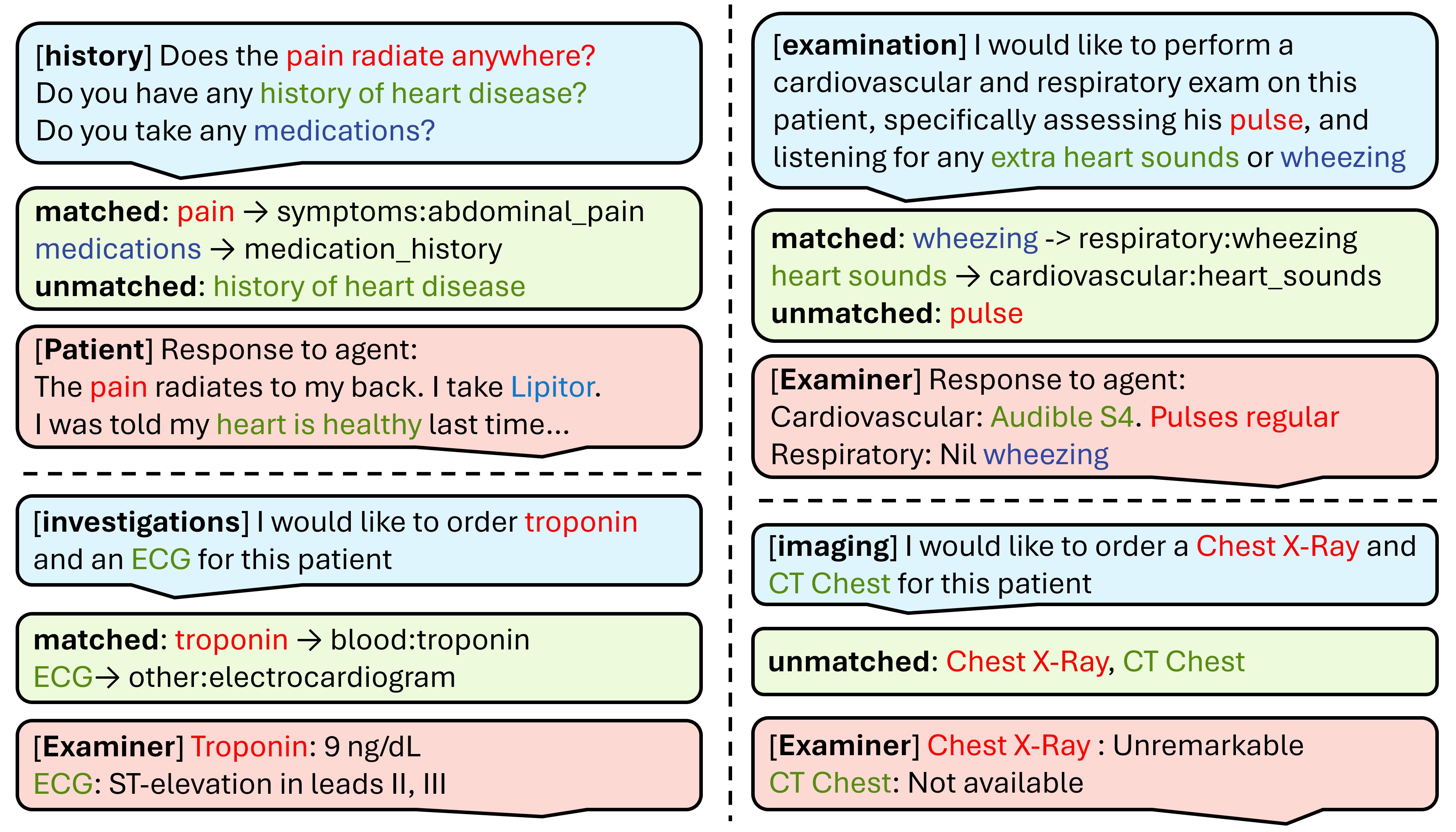}
    \caption{Example of how our Mapper module ($\mathbb{M}$, shown in green) maps specific keywords from free-text queries from agent (blue), across history ($\mathcal{H}$, top left), physical examination ($\mathcal{P}$, top right), lab investigations ($\mathcal{L}$, bottom left), and imaging ($\mathcal{I}$, bottom right). Each matched key from $\mathcal{K}_C$ has a corresponding value from $\mathcal{V}_C$, which is then processed by the Parser module ($\mathbb{P}$) into a free-text response (red). For any unmatched query terms, relevant negative findings are included for history and examination queries, and default normal values are provided for common investigations if not specified in the case (e.g., Chest X-Ray showing no acute abnormality).}
    \label{fig:mapper-example}
\end{figure}

We implemented two variants of the mapper: 1. A deterministic variant ($\mathbb{M}_D$) that maps free-text queries to relevant keys using cosine similarity embeddings, medical entity recognition, and domain-specific keyword matching. 2. An LLM-based variant ($\mathbb{M}_{LLM}$) that utilizes few-shot examples to interpret the semantic intent of queries. Our experiments use \texttt{gpt-4.1} with \texttt{temperature = 0} as the backbone for $\mathbb{M}_{LLM}$. Both approaches return a set of matched key-value pairs derived from $\mathcal{K}_C$ (and their associated values from $\mathcal{V}_C$) and a list of unmatched terms from the original query, thereby maintaining explicit boundaries regarding available information. Figure \ref{fig:mapper-example} illustrates this process with a working example.

The parser module, $\mathbb{P}$, then transforms the retrieved structured information (the matched key-value pairs from $\mathbb{M}$) into coherent natural language responses formatted according to medical documentation conventions. For history ($\mathcal{H}$) and physical examination ($\mathcal{P}$) findings, negative results (e.g., absence of a symptom) are explicitly returned if queried. Standardized laboratory values not specifically detailed in the source case $\mathcal{C}$ are returned as default normal values with appropriate reference ranges. Investigations not available in $\mathcal{C}$ are explicitly noted as "not available" to prevent information leakage or hallucination by the agent. Our framework also provides an optional LLM enhancement layer for $\mathbb{P}$ for history and physical examination outputs. This layer (using \texttt{gpt-4.1}) refines the natural language generation of responses, improving their fluency, while strictly adhering to the information boundaries established by the deterministic parsing logic. This ensures that responses remain factually accurate and grounded in $\mathcal{C}$, facilitating human review during evaluation.

\subsection{Mapper Module ($\mathbb{M}$}
\textbf{Deterministic Mapper ($\mathbb{M}_D$)} The deterministic variant, $\mathbb{M}_D$, translates free-text queries into structured information keys through a multi-stage process detailed in Algorithm \ref{alg:mapper_deterministic_algo}. For queries related to history ($\mathcal{H}$) and physical examination ($\mathcal{P}$), relevant keys from $\mathcal{K}_C$ are identified based on semantic similarity (computed using a dual-embedding approach with PubMedBERT and SNOMED-CT mediated similarity) between extracted query phrases and standardized medical terms. For laboratory ($\mathcal{L}$) and imaging ($\mathcal{I}$) requests, synonym dictionaries including terminology from medical coding systems (e.g., LOINC), common abbreviations, and predefined investigation panels are used to recognize diagnostic tests across various terminologies.

\begin{algorithm}[htb]
\caption{Process for deterministic mapper $\mathbb{M}_D$}
\label{alg:mapper_deterministic_algo}
    \begin{enumerate}[leftmargin=1.5em]
    \item Medical entities $\mathcal{V}_q$ are extracted from the input free-text query $\mathcal{Q}$ using the SpaCy biomedical model \texttt{en\_core\_sci\_md}.
    \item For history ($\mathcal{H}$) and physical examination ($\mathcal{P}$) queries:
        \begin{itemize}[leftmargin=0.5em]
            \item An embedding model based on PubMedBERT embeddings (\texttt{neuml/pubmedbert-base-embeddings}) is initialized, and indexed with $\mathcal{V}_C$, the set of textual descriptions associated with keys in $\mathcal{K}_C$.
            \item For each extracted query entity $v \in \mathcal{V}_q$:
            \begin{itemize}
                \item Direct semantic similarity, $\text{sim}(v, v_c)$, is computed between $v$ and each case-specific description $v_c \in \mathcal{V}_C$.
                \item SNOMED-CT mediated similarity is computed: $v$ is first mapped to a set of SNOMED-CT concepts $S_v$ where $\text{sim}(v, s) > \tau_S$ (with $\tau_S = 0.8$). Then, $\text{sim}(s, v_c)$ is computed for each $s \in S_v$ and $v_c \in \mathcal{V}_C$.
            \end{itemize}
            \item Query entities are matched to keys in $\mathcal{K}_C$ if either their direct or SNOMED-CT mediated similarity to a $v_c$ exceeds a predefined threshold $\tau_M = 0.6$.
        \end{itemize}
        This approach normalizes terminological variations (e.g., "shortness of breath" vs. "dyspnea") through medical concept alignment.

    \item For laboratory investigation ($\mathcal{L}$) and imaging ($\mathcal{I}$) queries:
        \begin{itemize}[leftmargin=0.5em]
            \item Comprehensive synonym dictionaries, $\mathbb{D}_{LAB}$ (for labs) and $\mathbb{D}_{IMG}$ (for imaging), are utilized to map common terms, LOINC codes, laboratory abbreviations, imaging modalities, and anatomical specifications to standardized keys in $\mathcal{K}_C$.
            \item For each query entity $v \in \mathcal{V}_q$, potentially relevant keys $\mathcal{K}_{relevant} = \mathbb{D}(v)$ are retrieved from the respective dictionaries.
            \item Matched keys are confirmed by intersecting with keys present in the current case: $\mathcal{K}_v = \mathcal{K}_{relevant} \cap \mathcal{K}_C$.
        \end{itemize}

    \item The mapper $\mathbb{M}_D$ returns:
        \begin{itemize}[leftmargin=0.5em]
            \item A set of matched pairs: $\{(v, k) \mid v \in \mathcal{V}_q, k \in \mathcal{K}_C, \text{and } k \text{ is matched to } v \text{ via steps 2 or 3}\}$.
            \item A set of unmatched query entities: $\mathcal{U}_q = \{v \mid v \in \mathcal{V}_q, v \text{ has no match in } \mathcal{K}_C\}$.
        \end{itemize}
    \end{enumerate}
\end{algorithm}

\textbf{LLM-based Mapper ($\mathbb{M}_{LLM}$)}
The LLM-based mapper, $\mathbb{M}_{LLM}$, employs an LLM (\texttt{gpt-4.1} with \texttt{temperature = 0} in our experiments) as its core component to semantically parse free-text clinical queries into structured information keys. For each query $\mathcal{Q}$ corresponding to an action category in $\{\mathcal{H}, \mathcal{P}, \mathcal{I}, \mathcal{L}\}$, an API call is made to the LLM. The prompt includes system instructions, few-shot examples demonstrating the mapping task, the user query $\mathcal{Q}$, and the set of available information keys $\mathcal{K}_C$ for the current clinical case. The LLM processes this input and returns a structured JSON response containing matched pairs $\{(v, k) \mid v \in \mathcal{V}_q, k \in \mathcal{K}_C\}$ and unmatched query entities $\mathcal{U}_q = \{v \mid v \in \mathcal{V}_q, v \text{ has no match in } \mathcal{K}_C\}$. This design allows $\mathbb{M}_{LLM}$ to handle complex or nuanced queries while preventing information leakage by grounding its output in $\mathcal{K}_C$.

To optimize $\mathbb{M}_{LLM}$, a calibration dataset of 100 clinical cases with human-annotated query-to-key mappings was used. Prompts were iteratively refined to maximize mapping accuracy against these gold-standard annotations, ensure output consistency across repeated identical requests, and improve robustness to variations in query phrasing. Examples included in system prompts were selected to cover common query patterns, including the correct handling of negative findings (e.g., "patient denies fever"), requests for normal values, and queries about unavailable information. This iterative refinement ensures that $\mathbb{M}_{LLM}$ maintains high precision in information retrieval while effectively managing the diversity of information requests encountered during agent interaction.

\begin{table}[htb]
    \centering
    \scriptsize 
    \setlength{\tabcolsep}{1.8pt} 
    \caption{Performance of the Deterministic ($\mathbb{M}_D$) and LLM-based ($\mathbb{M}_{LLM}$ with \texttt{gpt-4.1} backbone) Mapper variants. Precision (Pr) and Recall (Rc) were computed on a calibration set of 100 cases, using queries sampled from each of the evaluated agent models across the four information categories: $\mathcal{H}$ (History), $\mathcal{P}$ (Physical Exam), $\mathcal{I}$ (Imaging), and $\mathcal{L}$ (Labs).}
    \label{tab:mapper-validation}
    \begin{tabularx}{\columnwidth}{@{} L{0.9in} *{16}{Y} @{}} 
    \toprule
    \multirow{3}{*}{\textbf{Agent Model}} & 
    \multicolumn{8}{c}{\textbf{Deterministic Mapper ($\mathbb{M}_D$)}} &
    \multicolumn{8}{c}{\textbf{LLM-based Mapper ($\mathbb{M}_{LLM}$)}} \\
    \cmidrule(lr){2-9} \cmidrule(lr){10-17}

    & \multicolumn{2}{c}{$\mathcal{H}$} & \multicolumn{2}{c}{$\mathcal{P}$} & \multicolumn{2}{c}{$\mathcal{I}$} & \multicolumn{2}{c}{$\mathcal{L}$}
    & \multicolumn{2}{c}{$\mathcal{H}$} & \multicolumn{2}{c}{$\mathcal{P}$} & \multicolumn{2}{c}{$\mathcal{I}$} & \multicolumn{2}{c}{$\mathcal{L}$} \\
    \cmidrule(lr){2-3} \cmidrule(lr){4-5} \cmidrule(lr){6-7} \cmidrule(lr){8-9} 
    \cmidrule(lr){10-11} \cmidrule(lr){12-13} \cmidrule(lr){14-15} \cmidrule(lr){16-17}
    
    & Pr & Rc & Pr & Rc & Pr & Rc & Pr & Rc 
    & Pr & Rc & Pr & Rc & Pr & Rc & Pr & Rc \\
    \midrule

    DeepSeek-R1 & 0.76 & 0.64 & 0.71 & 0.87 & 0.98 & 0.87 & 0.82 & 0.84 & 0.84 & 0.94 & 0.86 & 0.91 & 1.00 & 0.97 & 0.92 & 0.90 \\
    o4-mini & 0.78 & 0.68 & 0.75 & 0.90 & 0.94 & 0.80 & 0.79 & 0.87 & 0.89 & 0.94 & 0.88 & 0.96 & 0.98 & 0.94 & 0.89 & 0.85 \\
    Llama-4 Maverick & 0.79 & 0.71 & 0.63 & 0.83 & 0.94 & 0.81 & 0.71 & 0.91 & 0.82 & 0.99 & 0.89 & 0.88 & 0.96 & 0.97 & 0.97 & 0.93 \\
    Grok 3 Mini Beta & 0.77 & 0.67 & 0.67 & 0.90 & 0.96 & 0.84 & 0.77 & 0.91 & 0.88 & 0.96 & 0.92 & 0.97 & 0.98 & 1.00 & 0.97 & 0.94 \\
    Qwen 3 (235b-a22b) & 0.82 & 0.64 & 0.75 & 0.95 & 0.94 & 0.78 & 0.75 & 0.88 & 0.84 & 0.94 & 0.87 & 0.95 & 0.98 & 0.97 & 0.95 & 0.88 \\
    Gemini 2.5 Pro & 0.77 & 0.59 & 0.72 & 0.81 & 0.97 & 0.78 & 0.80 & 0.87 & 0.83 & 0.94 & 0.90 & 0.92 & 1.00 & 0.96 & 0.94 & 0.83 \\
    \bottomrule
    \end{tabularx}
\end{table}

\textbf{Validation of Mapper Modules}
\label{ssec:mapper_validation}
To validate the performance and robustness of our mapper modules ($\mathbb{M}_D$, $\mathbb{M}_{LLM}$), we evaluated their precision (Pr) and recall (Rc) in mapping free-text queries to the correct structured information keys ($\mathcal{K}_C$). This validation was conducted on a calibration set of 100 clinical cases featuring human-annotated query-to-key mappings. Queries were sampled from those generated by each of the evaluated agent models across the four clinical categories ($\mathcal{H, P}\mathcal{, I, L}$). Specifically, for each agent model, 100 queries were sampled for each category to serve as input to $\mathbb{M}_D$ and $\mathbb{M}_{LLM}$.

Table \ref{tab:mapper-validation} outlines the result our validation. Overall, both mapper variants are largely consistent when processing queries generated by different agent models. This suggests that the mappers do not exhibit significant bias towards the query style of any particular agent architecture, or for $\mathbb{M}_{LLM}$, its base LLM model family (i.e. o4-mini vs. others). The $\mathbb{M}_{LLM}$ variant generally showed higher Pr/Rc, particularly for $\mathcal{H}$ and $\mathcal{P}$ queries. This was likely due to limitations on entities extraction process in $\mathbb{M}_{D}$, where it is unable to handle the richness and complexity in descriptive free-text queries. 

Furthermore, to confirm the operational determinism of $\mathbb{M}_{LLM}$ for identical inputs, we conducted an additional test where 100 distinct queries were each submitted 10 times to $\mathbb{M}_{LLM}$ (with a fixed case context). This experiment yielded an Intersection over Union (IoU) of $>$0.99 for the set of mapped keys and a variance of $<$0.01 in the mapped output across all repetitions. These findings confirm that $\mathbb{M}_{LLM}$, despite using a LLM to assist with information processing, produces highly deterministic and reliable mappings for identical queries within a given clinical case context. For the rest of our experiments, we use $\mathbb{M}_{LLM}$ due to its superior performance over $\mathbb{M}_{D}$

To assess inter-rater reliability, we conducted a post-hoc validation using two independent clinicians on 50 cases, with a weighted Cohen's kappa of 0.655 (moderate agreement). 

\subsection{Parser Module ($\mathbb{P}$)}
\label{ssec:parser_module}
After the Mapper module ($\mathbb{M}_D$ or $\mathbb{M}_{LLM}$) translates an agent's query into a set of structured (key, value) pairs corresponding to information available in $\mathcal{K}_C$, a Parser module ($\mathbb{P}$) transforms these retrieved clinical values into coherent natural language responses. For each action category ($\mathcal{H, P}\mathcal{, I, L}$), $\mathbb{P}$ formats the matched clinical information using pre-set templates designed to emulate standard medical documentation. $\mathbb{P}$ also maintains a record of previously processed keys within an interaction session to prevent redundant information retrieval and presentation.

For history ($\mathcal{H}$) and physical examination ($\mathcal{P}$) findings, unmatched requests are considered negative results, and are returned as relevant negatives if the information is not available in $\mathcal{C}$  (e.g., "Query: Does the patient have a fever? Response: Negative for fever."). For common investigations in $\mathcal{I, L}$ not specifically detailed in $\mathcal{C}$, default normal values are returned within predefined reference ranges. Uncommon lab investigations or imaging requests for data not available in $\mathcal{C}$ are explicitly reported as "not available" to prevent agent hallucination.

The framework also includes an optional LLM-based enhancement layer that can be applied to the output of $\mathbb{P}$ specifically for history and physical examination responses. This layer utilizes the same LLM as $\mathbb{M}_{LLM}$ (\texttt{gpt-4.1}). For history queries, it simulates patient responses in a first-person perspective, maintaining appropriate medical literacy. For physical examination queries, it formats findings as concise clinical notes organized by body system. This LLM enhancement improves the naturalness and readability of responses for human review, while strictly adhering to the information boundaries and factual content established by the core deterministic parsing logic.

\section{Evaluation Framework}
\label{app:evaluation_framework}

\textbf{Evaluated Models}
We evaluated several state-of-the-art foundation models, including Gemini 2.5 pro \cite{GoogleGemini2_5Pro2025}, DeepSeek-R1 \cite{DeepSeekAI2025DeepSeekR1}, o4-mini from OpenAI \cite{OpenAIO3O4miniSystemCard2025}, Llama-4 Maverick \cite{MetaLlama4Maverick2025}, Grok 3 mini beta \cite{xAI_Grok3mini_beta_2025}, and Qwen 3 (235b-a22b) \cite{AlibabaQwen3_235B_A22B_2025}. These models were selected based on their performance ranking on the Artificial Analysis Leaderboard \cite{ArtificialAnalysisLLMLeaderboard}. Each model was accessed through OpenRouter \cite{OpenRouter} and tested at \texttt{temperature = 0}.

\textbf{Evaluation Task}
Each structured clinical case $\mathcal{C}$ in our VivaBench benchmark consists of five components: history ($\mathcal{H}$), physical examination ($\mathcal{P}$), imaging investigations ($\mathcal{I}$), laboratory investigations ($\mathcal{L}$), and a set of ground truth diagnoses ($\mathcal{D}$) with accepted differentials ($\mathcal{D}'$).

Models interacted with these cases through a free-flowing dialogue system. They could request information from any category using specific actions (e.g., \texttt{history}, \texttt{examination}, \texttt{investigation}, \texttt{imaging}) and received responses synthesized from the structured case content. Models were instructed to first gather patient information through history-taking and physical examination, then provide a provisional diagnosis ($d_P$) before ordering any diagnostic tests. Subsequently, they were to order necessary investigations and, upon determining they had sufficient information, provide a final diagnosis ($d_D$). The complete system prompt detailing these instructions and available actions is provided in Algorithm \ref{alg:system_prompt_in_appendix}. 

We imposed a global turn limit of 20 interaction steps per case. Category-specific request limits were also enforced: 10 for history ($\mathcal{H}$), 5 for physical examination ($\mathcal{P}$), 3 for imaging ($\mathcal{I}$), and 3 for laboratory ($\mathcal{L}$) requests. Each information request counted as one step towards the global limit. If an agent reached a category-specific limit, it was instructed to proceed to the next phase of the diagnostic process. Failed requests due to formatting errors could be retried up to twice per original request; these retries counted towards the global action limit but not the category-specific limits.

Our evaluation assessed model performance at two key stages:
\begin{itemize}
    \item \textbf{Provisional Stage}: After the initial review of patient history and physical examination findings, but before ordering any investigations, the model provided one or more provisional diagnoses ($d_P$) with associated confidence values.
    \item \textbf{Final Stage}: After ordering and reviewing relevant diagnostic investigations, the model provided one or more final diagnoses ($d_D$) with updated confidence values.
\end{itemize}

For both provisional and final diagnoses, models were instructed to provide up to five possible diagnoses in a specified JSON format, each including: a free-text condition name, the corresponding ICD-10 name, the ICD-10 code, and a confidence score (0.0-1.0).

Additionally, we conducted a \textbf{Full Information} pre-test. In this setting, models were provided with the complete clinical picture (all information from $\mathcal{H}, \mathcal{P}, \mathcal{I}, \mathcal{L}$) upfront, bypassing the interactive information-gathering process. This served as a control to differentiate knowledge recall and synthesis capabilities from active information-seeking and sequential reasoning abilities.

\textbf{Evaluation Data and Metrics}
The complete output trace for each model interaction—including all queries, retrieved information (matched keys), provisional diagnoses, and final diagnoses—was preserved. All evaluated models achieved a first-pass success rate (i.e., completed the interaction without critical errors) of over 97\% on the VivaBench dataset. For subsequent metric computation, we used the intersection of successfully completed cases across all models, resulting in an evaluation set of 934 cases. The specific metrics computed for diagnostic accuracy, confidence calibration, information seeking efficiency, and diagnostic adaptation are detailed in Appendix \ref{app:evaluation_metrics}. Full experimentation logs are available upon request.

\textbf{Computational Cost}
The token usage and approximate costs associated with executing one full run of the VivaBench benchmark (evaluating one model over 934 cases) are summarized in Table \ref{tab:cost_summary_appendix}. On average, evaluating one model requires 26.4M tokens on average, with estimated costs ranging from \$11.9 to \$389 depending on the specific LLM's pricing at the time of experimentation. These cost figures can be considered the upper bound for each model, as we did not consider the price difference in input / output tokens, or account for potential cost reductions from caching mechanisms for repeated prompts or API calls in our calculations.

\begin{algorithm}
\caption{{System prompt for AI agent for evaluation task (Examples omitted)}}
\label{alg:system_prompt_in_appendix}

You are a primary care medical AI assistant. You are currently reviewing a patient. 
Your goal is to perform a full diagnostic workup for the patient, and find the underlying diagnosis to the patient's presentation. 

Workflow constraints:
\begin{enumerate}
    \item You must first gather patient information through history and examination before ordering any tests
    \item After reviewing the patient at bedside, you should provide a provisional diagnosis, before ordering any investigations
    \item Once you order any lab or imaging investigations, you can no longer gather additional history or perform examinations on the patient
    \item You can only perform one action at a time
    \item When you have sufficient information, you should provide a final diagnosis
\end{enumerate}

Available actions:
\begin{itemize}
    \item \texttt{history}: Interview the patient directly. Ask only 1-2 questions at a time to avoid overwhelming them. Assume average medical literacy.
    \item \texttt{examination}: Perform a physical examination. Specify exactly what examination you want to perform and what signs you're looking for.
    \item \texttt{diagnosis\_provisional}: Provide your provisional diagnosis given a clinical picture, after reviewing the patient but before ordering any investigations or imaging.
    \item \texttt{investigation}: Order any investigations that are not imaging. If you are ordering a laboratory test, specify which laboratory tests you are ordering, and specimen type if the laboratory test you are ordering is not serological. Bedside tests such as ECG, and other special tests, such as EEG, Pulmonary Function Tests etc., go here as well.
    \item \texttt{imaging}: Order medical imaging. Imaging modalities are strictly limited to imaging modalities that are performed by a radiologist, radiographer, or nuclear medicine physician, such as xray, ultrasound, CT, MRI, PET-scan etc. VQ scan also included here. Specify both the modality and anatomical region.
    \item \texttt{diagnosis\_final}: Provide your final diagnosis after completing your evaluation.
\end{itemize}

For diagnoses (both provisional and final):
\begin{itemize}
    \item Some patients might have multiple issues/diagnoses, or you may not be certain about this patient's diagnosis. You may list up to five possible diagnoses if there are multiple or if you are uncertain.
    \item For each diagnosis, provide the condition name, ICD-10 name, ICD-10 code, and your confidence (0.0-1.0) about the diagnosis. The condition name can be any descriptive text you choose, while the ICD-10 name needs to adhere to ICD-10 terminology.
    \item Confidence scores do not need to sum to 1.0
    \item Format as a list of dictionaries: \texttt{[{"condition": "free text name of the condition", "icd\_10\_name": "icd 10 name of the condition", "icd\_10": "icd code of the condition", "confidence": score}]}
    \item Remember to always give your provisional diagnosis before ordering any investigations
\end{itemize}
\end{algorithm}


\begin{table}[htb]
    \centering
    \footnotesize
    \caption{Approximate token usage and costs per model for a full VivaBench evaluation run (934 cases). "Per Case Tokens (k)" indicates the mean number of tokens (in thousands) used per case, including standard deviation. Costs are estimated based on OpenRouter pricing prevailing during experiments.}
    \label{tab:cost_summary_appendix} 
    \setlength{\tabcolsep}{4pt}
    \begin{tabular}{@{}l r r r r r@{}}
    \toprule
    \textbf{Model} & \textbf{Success (\%)} & \textbf{Total Tokens (M)} & \textbf{Per Case Tokens (k)} & \textbf{Rate (\$/M)} & \textbf{Est. Cost (\$)} \\
    \midrule
    Gemini 2.5 Pro    & 97.1 & 38.9 & 40.4 $\pm$ 8.3 & 10.00 & 389.00 \\
    DeepSeek-R1       & 98.6 & 20.0 & 20.5 $\pm$ 3.8 &  2.50 &  50.00 \\
    o4-mini           & 98.1 & 31.0 & 31.9 $\pm$ 9.2 &  4.40 & 136.40 \\
    Llama-4 Maverick  & 99.9 & 22.7 & 23.0 $\pm$ 4.6 &  0.60 &  13.60 \\
    Grok 3 Mini Beta  & 100.0& 23.8 & 24.1 $\pm$ 3.5 &  0.50 &  11.90 \\
    Qwen 3 (235b-a22b)& 100.0& 21.9 & 22.1 $\pm$ 4.0 &  0.60 &  13.10 \\
    \bottomrule
    \end{tabular}
\end{table}

\section{Evaluation Metrics} 
\label{app:evaluation_metrics}

\begin{table}[!htbp]
    \centering
    \footnotesize
    \caption{Expanded Top-k \textbf{Approximate} Diagnostic Accuracy across different evaluation stages: Provisional diagnosis ($d_P$), Final diagnosis ($d_D$ after interaction), and Full Information pre-test ($d_F$). These approximate accuracy metrics complement the exact accuracy results presented in Table \ref{tab:model_performance_metrics} in the main paper.}
    \label{tab:expanded_approximate_accuracy_metrics} 
    \setlength{\tabcolsep}{4pt} 
    \begin{tabularx}{\columnwidth}{@{} L{1.2in} *{9}{Y} @{}} 
    \toprule
    \multirow{2}{*}{\textbf{Model}} & 
    \multicolumn{3}{c}{\textbf{Provisional ($d_P$)}} &
    \multicolumn{3}{c}{\textbf{Final ($d_D$)}} &
    \multicolumn{3}{c}{\textbf{Full Information ($d_F$)}} \\
    \cmidrule(lr){2-4} \cmidrule(lr){5-7} \cmidrule(lr){8-10}
    & Top-1 & Top-3 & Top-5 & Top-1 & Top-3 & Top-5 & Top-1 & Top-3 & Top-5 \\
    \midrule
    Gemini 2.5 Pro    &  0.65  &  0.70  &  0.72  &  0.75  &  0.79  &  0.80  &  0.93  &  0.98  &  0.98 \\
    DeepSeek-R1       &  0.61  &  0.66  &  0.67  &  0.71  &  0.75  &  0.75  &  0.93  &  0.97  &  0.97 \\
    o4-mini           &  0.62  &  0.67  &  0.68  &  0.74  &  0.77  &  0.77  &  0.94  &  0.98  &  0.98 \\
    Llama-4 Maverick  &  0.59  &  0.64  &  0.64  &  0.70  &  0.71  &  0.71  &  0.91  &  0.96  &  0.96 \\
    Grok 3 Mini Beta  &  0.61  &  0.66  &  0.68  &  0.65  &  0.70  &  0.72  &  0.92  &  0.96  &  0.96 \\
    Qwen 3 (235b-a22b)&  0.56  &  0.60  &  0.61  &  0.68  &  0.72  &  0.72  &  0.90  &  0.96  &  0.96 \\
    \bottomrule
    \end{tabularx}
\end{table}


\subsection{Diagnostic Accuracy}
\label{ssec:metrics_diag_accuracy_appendix} 
We assessed diagnostic accuracy using several complementary metrics, focusing on the models' ability to identify correct diagnoses within their top-k predictions at different stages of the evaluation (Provisional $d_P$, Final $d_D$, and Full Information pre-test).

\textbf{Top-k Exact Accuracy}
We determined whether any of an agent's top-k predictions exactly matched a ground truth diagnosis in $\mathcal{D}$. A prediction was considered an exact match if its specified condition name or ICD-10 name/code precisely matched an entry in $\mathcal{D}$. For the Provisional ($d_P$) and Final ($d_D$) interactive stages, an additional criterion for an exact match was that the agent must have also ordered at least one relevant investigation supporting that specific diagnosis; this criterion does not apply to the Full Information stage. Exact accuracy results are presented in Table \ref{tab:model_performance_metrics} (main paper).

\textbf{Top-k Approximate Accuracy}
We also assessed whether any of an agent's top-k predictions approximately matched a ground truth diagnosis in $\mathcal{D}$ or an accepted differential diagnosis in $\mathcal{D}'$. A prediction was considered an approximate match if it met any of the following conditions:
\begin{itemize}
    \item It matched the ICD-10 code of any diagnosis in $\mathcal{D}$ at a broader hierarchical level (e.g., predicting I23 for a ground truth of I23.1).
    \item It matched the ICD-10 name or condition name of any accepted differential diagnosis listed in $\mathcal{D}'$.
    \item Its ICD-10 name or condition name exhibited high semantic similarity (cosine similarity $>0.8$) to any diagnosis in $\mathcal{D}$ or $\mathcal{D}'$.
\end{itemize}
For the Provisional ($d_P$) and Final ($d_D$) diagnostic stages, a prediction qualifying as an approximate match under these criteria also required the agent to have ordered at least one relevant investigation supporting that diagnosis or differential. This investigation ordering criterion does not apply to the Full Information ($d_F$ stage. Expanded Top-k Approximate Accuracy results are presented in Table \ref{tab:expanded_approximate_accuracy_metrics}.

\textbf{Multiple Diagnosis Handling}
For clinical cases in $\mathcal{C}$ with multiple ground truth diagnoses specified in $\mathcal{D}$, accuracy (both exact and approximate) was counted as a success (value of 1.0) if any one of the ground truth diagnoses was correctly identified by the agent within its top-k predictions, according to the respective criteria.

\subsection{Confidence Metrics}
\label{ssec:metrics_confidence_appendix} 
For each diagnosis item provided by an agent in its provisional ($d_P$) and final diagnosis ($d_D$) lists, a confidence score (0.0 to 1.0) was required.

\textbf{Raw Confidence Values}
We tracked the mean raw confidence scores across all diagnoses provided by an agent at the provisional stage ($C_P$ in Table \ref{tab:model_performance_metrics}) and final diagnosis stage ($C_D$ in Table \ref{tab:model_performance_metrics}).

\textbf{Confidence-Weighted Accuracy Score ($S_{\text{conf}}$)}
To evaluate how well an agent's confidence aligns with its diagnostic correctness, we first normalized the confidence scores ($c_j$) for all $N$ diagnoses proposed by the agent in a given stage for a case, such that $\sum_{j=1}^{N} c'_j = 1.0$, where $c'_j$ is the normalized confidence. We then calculated $S_{\text{conf}}$ as:
\[S_{\text{conf}} = \sum_{i \in E} c'_i + \sum_{i \in A} c'_i - \sum_{i \in U} c'_i\]
Where $c'_i$ is the normalized confidence for the $i$-th prediction, and $E, A, U$ are the sets of exact, approximate, and unmatched diagnostic predictions respectively. This score ranges from -1.0 to +1.0. Results for the provisional and final diagnostic stages are presented as $S_P$ and $S_D$ in Table \ref{tab:model_performance_metrics} of main paper respectively.

\textbf{Confidence-Accuracy Matrix}
To explore the relationship between diagnostic performance and confidence calibration, we mapped models' confidence against accuracy using kernel density estimation (KDE) distributions (Figure 5, main paper). For each clinical case, a model's mean raw confidence score (y-axis) was plotted against its composite diagnostic accuracy (x-axis; defined as average of Top-k exact and approximate accuracies across $k \in \{1, ..., 5\}$). KDE contours (bandwidth 0.7, 50\% density mass) were calculated to visualize case distribution per model. The matrix quadrants (Confidently Accurate, Overconfidently Wrong, Underconfident but Accurate, Appropriately Uncertain) denote distinct patterns in how different models calibrate uncertainty against their knowledge within our simulated clinical setting. 

\subsection{Information Seeking Efficiency}
\label{ssec:metrics_info_seeking_appendix} 
We evaluated how models strategically gather clinical information by measuring \textit{Information seeking efficiency}, which we define as  the ability to selectively request relevant history ($\mathcal{H}$), physical examination ($\mathcal{P}$), and diagnostic investigations ($\mathcal{I, L}$) that contribute to an accurate diagnosis. To measure this, we tracked binarised metrics of matched information keys in our evaluation:
\begin{equation} \label{eq:precision_recall_appendix} 
\text{Precision} = \frac{|K_{\text{requested}} \cap K_{\text{relevant}}|}{|K_{\text{requested}}|} \qquad \text{Recall} = \frac{|K_{\text{requested}} \cap K_{\text{relevant}}|}{|K_{\text{relevant}}|}
\end{equation}
Where $K_{\text{requested}}$ is the set of clinical information elements requested by the model, and $K_{\text{relevant}}$ is the set expert-annotated as relevant for diagnosis. Metrics were averaged across cases.
Calculations used four configurations (Table \ref{tab:info_seeking_config_appendix_e}), based on information stage (Review vs. Investigation) and diagnostic scope (Matched vs. Full). In general, "Matched Scope" assesses targeted inquiry; "Full Scope" evaluates thoroughness. Precision reflects inquiry focus, while Recall indicates completeness.

\begin{table}[htb] 
\centering
\caption{Configurations for Information Seeking Efficiency Metrics.}
\label{tab:info_seeking_config_appendix_e} 
\begin{tabular}{|l|p{0.4\columnwidth}|p{0.4\columnwidth}|} 
\hline
& \textbf{Matched Scope} & \textbf{Full Scope} \\
& (Targeted to proposed diagnoses) & (All diagnosis-relevant information) \\
\hline
\textbf{Review Stage} & Precision and recall of history ($\mathcal{H}$) and physical examination ($\mathcal{P}$) items relevant \textit{only} to the diagnoses actively proposed by the agent. & Precision and recall of \textit{all} potentially relevant history ($\mathcal{H}$) and physical examination ($\mathcal{P}$) items for the case, regardless of the agent's current diagnostic hypotheses. \\
\hline
\textbf{Investigation Stage} & Precision and recall of imaging ($\mathcal{I}$) and laboratory ($\mathcal{L}$) investigations relevant \textit{only} to the diagnoses actively proposed by the agent. & Precision and recall of \textit{all} potentially relevant imaging ($\mathcal{I}$) and laboratory ($\mathcal{L}$) investigations for the case, regardless of the agent's current diagnostic hypotheses. \\
\hline
\end{tabular}
\end{table}

\subsection{Diagnostic Evolution Metrics}
\label{ssec:metrics_diag_evolution_appendix} 
To further scrutinize the reasoning process within the evaluated models, we analyzed how their diagnostic hypotheses and associated confidence levels evolved between the provisional ($d_p$) and final ($d_f$) stages of the interaction. This examination aims to determine whether an agent's capacity to adapt its diagnostic judgment ("change its mind") and recalibrate its confidence in response to new information (from investigations) correlates with material improvements in diagnostic performance.

\textbf{Outcome Metrics}
To quantify the net change in diagnostic performance from the provisional to the final stage, we measured the following outcome metrics:
\begin{itemize} 
    \item \textbf{Exact Accuracy Change}: The change in Top-1 Exact Accuracy (Final $d_f$ - Provisional $d_p$). (Corresponds to "Acc Chg." in Table \ref{tab:correlation_analysis}).
    \item \textbf{Approximate Accuracy Change}: The change in Top-1 Approximate Accuracy (Final $d_f$ - Provisional $d_p$). (Corresponds to "Apr. Acc Chg." in Table \ref{tab:correlation_analysis}).
    \item \textbf{$S_{\text{conf}}$ Change}: The change in the Confidence-Weighted Accuracy Score ($S_{\text{conf}}$) (Final $d_f$ - Provisional $d_p$). (Corresponds to "$S_{\text{conf}}$ Chg." in Table \ref{tab:correlation_analysis}).
\end{itemize}

\textbf{Variable Metrics}
To characterize the specific adjustments made by models to their diagnostic hypotheses, we tracked the following variable metrics:
\begin{itemize} 
    \item \textbf{Diagnoses Added/Removed/Maintained}: The counts of unique diagnoses that were added to, removed from, or maintained in the agent's list of top-5 hypotheses when transitioning from the provisional to the final stage. (Corresponds to "Diag. Added", "Diag. Removed", "Diag. Maintained" in Table \ref{tab:correlation_analysis}).
    \item \textbf{Confidence Delta}: The overall change in mean raw confidence across all proposed diagnoses, calculated as (Mean Raw Confidence at Final Stage, $C_D$) - (Mean Raw Confidence at Provisional Stage, $C_P$). (Corresponds to "Conf. Delta ($C_D-C_P$)" in Table \ref{tab:correlation_analysis}).
    \item \textbf{Confidence Shift}: For diagnoses that were maintained in the agent's list from the provisional to the final stage, this is the mean of the difference in their confidence scores ($c_{f,j} - c_{p,j}$ for each maintained diagnosis $j$). (Corresponds to "Conf. Shift" in Table \ref{tab:correlation_analysis}).
    \item \textbf{Confidence Shift Magnitude}: For diagnoses maintained between stages, this is the mean of the absolute difference in their confidence scores ($|c_{f,j} - c_{p,j}|$ for each maintained diagnosis $j$). (Corresponds to "Conf. Shift Mag." in Table \ref{tab:correlation_analysis}).
\end{itemize}

\textbf{Correlation Analysis} 
\label{ssec:metrics_correlation_appendix} 
We analyzed the statistical relationships between these diagnostic adaptation variable metrics and the observed outcome metrics. The objective was to identify whether specific patterns of adjusting diagnostic hypotheses (e.g., adding/removing diagnoses, magnitude of confidence shifts) systematically correlate with improvements in final diagnostic accuracy or confidence calibration. For categorical outcome variables (e.g., whether exact accuracy improved), Welch's t-test (for two groups) or one-way ANOVA (for more than two groups) was applied to the variable metric distributions. For continuous outcome variables (e.g., the numerical change in $S_{\text{conf}}$), Pearson correlation coefficients were computed between the outcome and variable metrics.
The results of this correlation analysis are presented and discussed in the main paper (Table \ref{tab:correlation_analysis}). Broadly, these findings suggest that more effective diagnostic performance is often associated with a greater willingness to adapt initial hypotheses and recalibrate confidence in light of new evidence from diagnostic investigations.


\section{Qualitative Analysis}\label{app:qualitative}
Below, we include several examination traces, including agent reasoning, action, and query, as examples of potential failure modes we observed in our evaluation. Any mentioned relevant diagnoses are highlighted in purple, and specific sections that reflect the issues are highlighted in red. 
\lstdefinelanguage{tracelang}{
    keywords=[1]{Agent}, 
    keywords=[2]{Patient, Examiner, Stem},    
    keywords=[3]{Diagnosis}, 
    sensitive=true,
}
\definecolor{usergreen}{RGB}{0,128,0}

\textbf{Premature Diagnostic Closure} In a case involving a 28-year-old male presenting with epigastric pain, the agent correctly identified two potential diagnoses from clinical review, pancreatitis (confidence score 0.7) and peptic ulcer disease (confidence score 0.2), and ordered the appropriate tests for pancreatitis. Laboratory results revealed elevated amylase and lipase, which are positive for pancreatitis. This prompted the agent to conclude its journey, and diagnosed the pancreatitis without further investigations. However, the agent failed to pursue testing for the underlying etiology. In this case, it was due to a stricture from a duodenal ulcer, which could have been identified through a H. pylori testing, which is a routine test for any patients with suspected peptic ulcer\cite{fashner2015diagnosis}. This reflects satisfaction-of-search bias, where identification of one diagnosis prematurely terminates the diagnostic process, potentially missing causative or concomitant conditions.

\lstset{
    language=tracelang,
    basicstyle=\ttfamily\small,
    keywordstyle=[1]\color{blue}\bfseries,    
    keywordstyle=[2]\color{usergreen}\bfseries,   
    keywordstyle=[3]\color{purple}\bfseries,     
    backgroundcolor=\color[RGB]{245,245,244},
    breaklines=true,
    showstringspaces=false,
    columns=fixed,  
    keepspaces=false,       
    basewidth={0.5em, 0.05em}, 
    frame=single,
    framesep=5pt,
    framerule=0.4pt,
    breakindent=0pt,
    tabsize=1,
    numbers=none,
    xleftmargin=0pt,      
    moredelim=**[s][\bfseries]{[}{]}, 
    moredelim=**[is][\color{red}\bfseries]{@}{@}, 
    moredelim=**[is][\color{purple}\bfseries]{|}{|}, 
}
\begin{lstlisting}
Examiner: Clinical case stem: 28 year old male presenting with epigastric pain.
Vitals: Heart Rate: 92. Blood Pressure: 122/78. Temperature: 37.1. Respiratory rate: 16. Oxygen saturation: 99.

Please review and diagnose the patient.
Agent: The patient is a young adult male presenting with epigastric pain and stable vital signs. To narrow down the differential diagnosis, I need more information about the characteristics of the pain, such as onset, duration, quality, and any aggravating or relieving factors.
[history] Can you describe the epigastric pain? When did it start, and is it constant or does it come and go?
Examiner: The pain started suddenly earlier today. It's in the upper middle part of my abdomen, and it's been severe and constant since it began. It doesn't really come and go - it's just always there. Nothing seems to make it better or worse.
Agent: Sudden, severe, constant epigastric pain in a young male could be due to several causes, including peptic ulcer disease (with or without perforation), acute pancreatitis, biliary pathology, or even less common causes like myocardial ischemia or referred pain. I need to clarify associated symptoms to narrow the differential.
[history] Have you experienced any nausea, vomiting, fever, or changes in your bowel movements? Have you noticed any blood in your vomit or stool?
Examiner: I have been feeling nauseous, but I haven't had any vomiting or fever. My bowel movements haven't changed, and I haven't noticed any blood in my vomit or stool.
Agent: I need to further characterize the pain and assess for risk factors or associated symptoms that could help narrow the differential. I want to know if the pain radiates anywhere and if the patient has any history of similar pain, alcohol use, or known medical conditions.
[history] Does the pain radiate to your back, chest, or shoulders? Have you ever had similar pain before? Do you drink alcohol or have any known medical conditions such as ulcers or gallstones?
Examiner: The pain does radiate to my back. I've never had similar pain before. I drink alcohol occasionally - about two cans of beer twice a week. I don't have any known medical conditions, and I've never had ulcers or gallstones as far as I can tell.
Agent: The sudden onset of severe, constant epigastric pain radiating to the back raises concern for acute pancreatitis, but other causes such as peptic ulcer disease (including perforation), biliary pathology, or even less likely cardiac causes should be considered. I need to examine the abdomen for tenderness, guarding, rebound, and any palpable masses to further narrow the differential.
[examination] I want to perform an abdominal examination, specifically checking for tenderness, guarding, rebound tenderness, palpable masses, and any signs of peritonism.
Examiner: Abdomen:
Tenderness: Epigastric tenderness with mild guarding.
Guarding: Mild guarding present.
Rebound tenderness: No rebound tenderness.
Palpable masses: No palpable masses.
Peritonism: No signs of peritonism.
Agent: The patient is a young male with sudden-onset, severe, constant epigastric pain radiating to the back, associated with nausea but no vomiting, fever, or GI bleeding. Examination reveals epigastric tenderness with mild guarding but no peritonism or masses. The most likely causes include |acute pancreatitis|, |peptic ulcer disease (possibly with early perforation or severe gastritis)|, and less likely biliary or cardiac causes given the location and radiation. The absence of peritonism and stable vitals make perforation less likely, but acute pancreatitis is a strong consideration given the pain's character and radiation.
[diagnosis provisional] |(condition: Acute pancreatitis, confidence: 0.7), (condition: Peptic ulcer disease, confidence: 0.2)|, (condition: Gastritis, confidence: 0.1)
Examiner: Thank you. Please proceed to imaging and lab investigations.
Agent: Acute pancreatitis is the leading provisional diagnosis given the sudden severe epigastric pain radiating to the back, nausea, and epigastric tenderness with mild guarding. To confirm this and assess severity, I will order serum amylase and lipase, as well as liver function tests to rule out biliary involvement. Additionally, an abdominal ultrasound is indicated to assess for gallstones or pancreatic abnormalities.
[investigation] Order serum amylase, serum lipase, and liver function tests (ALT, AST, ALP, GGT, bilirubin).
Examiner: Blood:

@Amylase: 1145.0 U/L@
@Lipase: 1103.0 U/L@
Alanine Aminotransferase (ALT): 45.0 IU/L
Aspartate Aminotransferase (AST): 43.0 IU/L
Alkaline Phosphatase (ALP): 97.0 IU/L
Gamma-Glutamyl Transferase (GGT): 149.0 U/L
Total Bilirubin: 0.64 mg/dL
Agent: The patient's clinical presentation of sudden severe epigastric pain radiating to the back, along with significantly elevated serum amylase and lipase, is highly consistent with @acute pancreatitis@. Liver function tests are only mildly elevated, and there is no evidence of biliary obstruction or hepatitis. Other differentials are much less likely given the lab findings and clinical picture.
[diagnosis final] (condition: Acute pancreatitis, confidence: 1.0)
Diagnosis: Acute pancreatitis secondary to duodenal ulcer-induced ampullary obstruction, @acquired duodenal stenosis from chronic duodenal ulcer@
\end{lstlisting}

\textbf{Inadequate investigations} In a case of cerebral infarct, the agent was presented with a 63-year-old male with facial droop, numbness, and significantly elevated blood pressure (212/123 mmHg). The agent correctly suspected a pontine infarct in its provisional diagnosis from the clinical signs alone, but only ordered a non-contrast CT scan, which had low sensitivity (approximately 41\%) of CT imaging for posterior fossa lesions such as pontine strokes\cite{hwang2012ComparativeSensitivityComputed}. The imaging only showed chronic ischemic changes, and the agent diagnosed the patient with a transient ischemic attack (TIA). While a non-contrast CT scan was first-line for stroke patients, this case illustrates how models follow procedural pathways without appropriately weighing test limitations against clinical presentation.

\lstset{
    language=tracelang,
    basicstyle=\ttfamily\small,
    keywordstyle=[1]\color{blue}\bfseries,    
    keywordstyle=[2]\color{usergreen}\bfseries,   
    keywordstyle=[3]\color{purple}\bfseries,     
    backgroundcolor=\color[RGB]{245,245,244},
    breaklines=true,
    showstringspaces=false,
    columns=fixed,  
    keepspaces=false,       
    basewidth={0.5em, 0.05em}, 
    frame=single,
    framesep=5pt,
    framerule=0.4pt,
    breakindent=0pt,
    tabsize=1,
    numbers=none,
    xleftmargin=0pt,      
    moredelim=**[s][\bfseries]{[}{]}, 
    moredelim=**[is][\color{red}\bfseries]{@}{@}, 
    moredelim=**[is][\color{purple}\bfseries]{|}{|}, 
}
\begin{lstlisting} 
Stem: 63/M presenting with facial droop and other non-specific neurological symptoms. HR 98, BP 212/123 mmHg. Please review and diagnose the patient. 
Agent: The patient presents with facial droop and neurological symptoms alongside significantly elevated blood pressure, suggesting a potential stroke or hypertensive emergency. I need to gather more information about the patient's symptoms and history to narrow down the differential diagnosis. 
[history] Can you describe the facial droop and other neurological symptoms you're experiencing? When did they start? 
Patient: It started suddenly about three hours ago. At first, it was on the right side of my face, but that went away by the time I got here. Then, while I was waiting, I felt some weakness on the left side across my whole face. I also felt a bit numb on the left side of my body, but it shifted to the right face and arm afterwards. 
Agent: The migratory nature of the patient's symptoms is concerning for a transient ischemic attack (TIA) or evolving stroke. The significantly elevated blood pressure could be either a cause or consequence of cerebrovascular pathology. I should enquire other associated symptoms. 
[history] Have you experienced any headache, vision changes, dizziness, confusion, or trouble with balance or coordination?
Patient: I also have some difficulty swallowing, and my wife said my speech was a bit slurred. However, I don't feel it now. 
Agent: The patient's symptoms of sudden onset facial droop, shifting numbness, difficulty swallowing, and slurred speech suggest a neurological condition. The fluctuating nature of the symptoms could indicate a complex or dynamic pathology. I need to examine the patient to assess neurological deficits and look for signs that could help narrow down the diagnosis. 
[examination] I want to perform a neurological examination to assess cranial nerve function, muscle strength, sensation, and reflexes, looking for signs of stroke, demyelinating disease, or other neurological conditions. 
Examiner: Neurological: Cranial nerves grossly intact, no facial asymmetry currently visible. Muscle strength normal in all limbs (5/5 bilaterally). Sensation decreased to pinprick and cold temperature on left side, particularly in the left upper and lower extremities, consistent with the patient's reported symptoms. Reflexes 2+ and symmetrical throughout. Gait appears normal with no ataxia. Coordination intact on finger-to-nose and heel-to-shin testing. No pronator drift observed. NIH Stroke Scale score 1 for mild left-sided sensory deficit.
Agent: This patient presents with acute onset left facial paralysis (upper and lower face), left facial sensory loss, dysarthria, left-sided cerebellar signs, and a severe headache, with a history of transient left then right-sided numbness. There is no limb weakness or sensory loss in the limbs. The blood pressure is severely elevated. The findings localize to the left lateral pons or medulla, suggesting a brainstem syndrome, most likely a lateral pontine (AICA) or lateral medullary (Wallenberg) syndrome. The severe hypertension raises concern for hypertensive emergency with possible brainstem ischemia or hemorrhage. Other differentials include demyelinating disease or a mass lesion, but the acute presentation and vascular risk factors make stroke most likely."
[provisional diagnosis] |(condition: Acute ischemic stroke, left lateral pontine or medullary syndrome, confidence: 0.7)|, (condition: Hypertensive emergency with possible brainstem involvement, confidence: 0.4), (condition: Intracerebral hemorrhage , confidence: 0.2), (condition: Multiple sclerosis, confidence: 0.1)
Examiner: Thank you. Please proceed to imaging and lab investigations.
Agent: The patient's presentation is highly suggestive of an acute brainstem event (ischemic or hemorrhagic stroke). An urgent non-contrast CT head is needed to rule out hemorrhage, and MRI brain with diffusion-weighted imaging is ideal for detecting acute ischemia. Given the hypertensive emergency, basic labs are also warranted to assess end-organ damage.
[imaging] Order a @non-contrast CT scan@ of the head to assess for acute hemorrhage or large infarct.
Examiner: == CT BRAIN ==
Chronic periventricular ischemic changes without evidence of acute ischemia or hemorrhage
Agent: @Imaging does not show acute infarct or hemorrhage@, but the patient has focal neurological deficits (left facial paralysis, left facial sensory loss, left-sided cerebellar signs, dysarthria) and hypertensive emergency. The findings are most consistent with an acute brainstem (pontine) syndrome, possibly a @transient ischemic attack (TIA) or a small vessel (lacunar) event not visible on imaging@. Demyelinating disease is less likely given the age and risk factors, but cannot be fully excluded. The chronic ischemic changes support a vascular etiology.'
[investigation] I would like to check this patient's lipid profile, HbA1c, and a thrombophilia screen to assess for risk factors for stroke and potential underlying causes.
Examiner: Investigation Results
Total cholesterol: 245 mg/dL 
LDL cholesterol: 168 mg/dL
HDL cholesterol: 38 mg/dL
Triglycerides: 195 mg/dL
HbA1c: 6.8%
Thrombophilia screen: Pending
Agent: The patient has focal neurological deficits (left facial paralysis, left facial sensory loss, dysarthria, left-sided cerebellar signs) with a history of fluctuating symptoms and severe hypertension. Imaging shows only chronic ischemic changes, with no acute infarct or hemorrhage. The clinical picture is most consistent with a brainstem (pontine) transient ischemic attack (TIA) or a small vessel (lacunar) syndrome, likely related to hypertensive small vessel disease. There is no evidence of acute stroke or hemorrhage on imaging, and labs are unremarkable. The most likely diagnosis is a hypertensive brainstem TIA, but a small vessel lacunar syndrome remains possible. Demyelinating disease is less likely given the age and lack of other features.
[final diagnosis] @(condition: Transient ischemic attack (TIA), brainstem, confidence: 0.7)@, (condition: Hypertensive encephalopathy (with small vessel disease), confidence: 0.5), (condition: Lacunar syndrome (chronic small vessel ischemic disease), confidence: 0.3)
Diagnosis: Acute medial pontomedullary junction infarction
\end{lstlisting}

\textbf{Inappropriate Hypothesis Generation} When presented with a 4-month-old infant with feeding difficulties, one agent inappropriately prioritized heart failure in its differential diagnosis, stating: "To differentiate potential heart failure from respiratory pathology, I need more detail on feeding tolerance, sweating, cyanosis, and progression of breathing difficulty." This lead to subsequent ordering of inappropriate tests that are routine for heart failure, while failing to investigate other possible causes of the patient's presentation. This demonstrates a fundamental misalignment between the agent's reasoning and patient-specific clinical probabilities, suggesting limitations in the model's ability to appropriately contextualize presentations within pediatric epidemiology.

\lstset{
    language=tracelang,
    basicstyle=\ttfamily\small,
    keywordstyle=[1]\color{blue}\bfseries,    
    keywordstyle=[2]\color{usergreen}\bfseries,   
    keywordstyle=[3]\color{purple}\bfseries,     
    backgroundcolor=\color[RGB]{245,245,244},
    breaklines=true,
    showstringspaces=false,
    columns=fixed,  
    keepspaces=false,       
    basewidth={0.5em, 0.05em}, 
    frame=single,
    framesep=5pt,
    framerule=0.4pt,
    breakindent=0pt,
    tabsize=1,
    numbers=none,
    xleftmargin=0pt,      
    moredelim=**[s][\bfseries]{[}{]}, 
    moredelim=**[is][\color{red}\bfseries]{@}{@}, 
    moredelim=**[is][\color{purple}\bfseries]{|}{|}, 
}
\begin{lstlisting}
Stem: 4 week old female presenting with poor feeding and trouble breathing.
Temperature 37.5 C, HR 172 bpm, BP 90/50 mmHg, RR 32/min, O2 sat 100%
Please review and diagnose the patient.
Agent: To differentiate potential heart failure from respiratory pathology, I need more detail on feeding tolerance, sweating, cyanosis, and progression of breathing difficulty.
[history] How many feeds does she take per day and how much volume each time? Does she sweat or turn blue when feeding or crying?
Examiner: She's been feeding a lot less than usual - maybe just a few times a day, and she only takes a small amount each time, not finishing her usual bottles. She doesn't sweat or turn blue when she's feeding or crying.
Agent: We need to characterize the breathing difficulty - whether it's tachypnea, retractions, grunting, apneic spells, or upper airway issues - to narrow the differential.
[history] Can you describe her trouble breathing? Does she have rapid breathing, grunting, nasal flaring, chest retractions, coughing, or any pauses in breathing (apnea)?
Examiner: She seems to be having trouble breathing, but I haven't noticed her breathing fast, grunting, flaring her nostrils, pulling in at her chest, coughing, or having any pauses where she stops breathing.
Agent: Poor feeding, tachycardia, and reported breathing difficulty raise concern for cardiac pathology or @early heart failure@. I need to check for murmurs, gallops, and hepatomegaly.
[examination] I want to listen to this patient's heart sounds for any murmurs or gallop rhythms and palpate the abdomen for hepatomegaly.
Examiner: Cardiovascular: Heart sounds regular, no murmur or gallop rhythm.
Abdominal: No hepatomegaly.
Agent: The caregiver reports breathing difficulty despite no overt distress; I need to assess the lungs for any subtle abnormalities.
[examination] I want to auscultate the patient's lungs for any crackles, wheezing, or diminished breath sounds, and inspect the chest for symmetry and any deformity.
Examiner: Respiratory: Bilateral crackles on auscultation. No wheezing. Breath sounds not diminished. Chest inspection normal, no asymmetry or deformity.
Agent: The infant's poor feeding, tachycardia, and bilateral crackles without primary respiratory distress signs are most consistent with @early congestive heart failure@ from a congenital cardiac lesion, with pneumonia less likely.
[diagnosis provisional] (condition: @Congestive heart failure@, likely secondary to congenital heart defect, confidence: 0.7), (condition: Ventricular septal defect, confidence: 0.2), (condition: Pneumonia, unspecified organism, confidence: 0.1)
Examiner: Thank you. Please proceed to imaging and lab investigations.
Agent: A chest X-ray will help assess heart size, pulmonary vascular markings, and evidence of pulmonary edema or structural abnormalities suggestive of a congenital heart lesion.
[imaging] I would like to order a Chest X-Ray (PA and lateral)
Examiner: == X-RAY CHEST ==
Chest distension without focal infection

Agent: To evaluate for heart failure severity, rule out infection, and assess for possible myocarditis, I will order blood tests including markers of inflammation and cardiac stress plus cultures and a viral panel.
[investigation] Complete Blood Count; C-Reactive Protein; @B-type Natriuretic Peptide (serum)@; Blood cultures; Viral respiratory PCR panel (nasopharyngeal swab)
Examiner: Blood:
- White Blood Cell Count: 9.2 -> 18.4 x 10^9/L  - on admission, after 48 hours
- Hemoglobin: 10.9 -> 10.8 g/dL  - on admission, after 48 hours
- Platelet Count: 553 -> 734 x 10^9/L  - on admission, after 48 hours
- C-Reactive Protein: <5 mg/L  - on admission and after 48 hours
- Mean corpuscular volume: Normal
- B-type natriuretic peptide: Normal
Microbiology:
- Blood Culture, Aerobic: negative
- Blood Culture, Anaerobic: negative
- Respiratory Viral Panel: negative  - including RSV and influenza

Agent: The presentation of poor feeding, respiratory crackles, chest distension on X-ray without infection, and signs of pulmonary overcirculation raise suspicion for an underlying structural cardiac abnormality @despite absence of a murmur@; echocardiography is the next best step to evaluate for congenital heart disease or cardiomyopathy.
[imaging] I would like to order a transthoracic echocardiogram
Examiner: Echocardiogram not available.

Agent: An electrocardiogram can reveal signs of cardiomyopathy or structural heart disease (e.g., chamber enlargement, conduction abnormalities) when echocardiography is unavailable.
[investigation] Electrocardiogram (ECG)
Examiner: Other:
- Electrocardiogram: Normal

Agent: The infant's presentation of poor feeding, tachycardia, bilateral pulmonary crackles on exam, chest distension on X-ray, and absence of infection or primary pulmonary disease is @most consistent with congestive heart failure@ from an underlying congenital cardiac lesion. Without echocardiography, the precise defect is uncertain but a large ventricular septal defect or total anomalous pulmonary venous return remain possible.
[diagnosis final] (condition: Congestive heart failure, confidence: 0.9), (condition: Ventricular septal defect, confidence: 0.4), (condition: Total anomalous pulmonary venous connection, confidence: 0.2)
\end{lstlisting}